\documentclass[10pt,twocolumn,letterpaper]{article}

\usepackage{iccv}
\usepackage{times}
\usepackage{epsfig}
\usepackage{graphicx}
\usepackage{amsmath}
\usepackage{amssymb}
\usepackage[accsupp]{axessibility}  

\usepackage{tabularx}
\usepackage{multirow}
\usepackage{booktabs}
\usepackage[font=small]{subcaption}
\usepackage{arydshln}
\usepackage{wrapfig}
\usepackage{url}            
\usepackage[font=small]{caption}
\usepackage{lipsum}  
\usepackage[bottom]{footmisc}
\usepackage{enumitem}
\setlist{nosep}

\makeatletter
\def\thickhline{%
	\noalign{\ifnum0=`}\fi\hrule \@height \thickarrayrulewidth \futurelet
	\reserved@a\@xthickhline}
\def\@xthickhline{\ifx\reserved@a\thickhline
	\vskip\doublerulesep
	\vskip-\thickarrayrulewidth
	\fi
	\ifnum0=`{\fi}}
\makeatother

\newlength{\thickarrayrulewidth}
\usepackage{pgfplots}
\usepackage{pgfplotstable}

\usepackage{tikz}
\usepackage{tikz-3dplot}
\usetikzlibrary{calc}
\usetikzlibrary{decorations.text}
\usetikzlibrary{external}
\usetikzlibrary{fit}
\usetikzlibrary{intersections}
\usetikzlibrary{positioning}
\usetikzlibrary{shapes}
\usepgfplotslibrary{colorbrewer}
\usepgfplotslibrary{colormaps}
\usepgfplotslibrary{groupplots}
\usepgfplotslibrary{polar}
\usepgfplotslibrary{fillbetween}

\newcommand\blfootnote[1]{%
	\begingroup
	\renewcommand\thefootnote{}\footnote{#1}%
	\addtocounter{footnote}{-1}%
	\endgroup
}

\usepackage[pagebackref=true,breaklinks=true,letterpaper=true,colorlinks,bookmarks=false]{hyperref}

\usepackage[capitalize]{cleveref}
\crefname{section}{Sec.}{Secs.}
\Crefname{section}{Section}{Sections}
\Crefname{table}{Table}{Tables}
\crefname{table}{Tab.}{Tabs.}

\newcommand{\secref}[1]{Section~\ref{#1}}

\newcommand{\figref}[1]{Fig.~\ref{#1}}
\newcommand{\eqnref}[1]{Eq.~\ref{#1}}
\newcommand{\tabref}[1]{Table~\ref{#1}}

\newcommand{\figreftwo}[2]{Figs.~\ref{#1} and~\ref{#2}}

\iccvfinalcopy 

\def\iccvPaperID{10716} 

\ificcvfinal\fi

\begin{document}

\title{Taming Contrast Maximization for Learning\\ Sequential, Low-latency, Event-based Optical Flow}

\author{Federico Paredes-Vall\'es$^{1,2}$\hspace{16pt}Kirk Y. W. Scheper$^2$\hspace{16pt}Christope De Wagter$^1$\hspace{16pt}Guido C. H. E. de Croon$^1$\vspace{5pt}\\
   $^1$ Micro Air Vehicle Laboratory, Delft University of Technology\\
   $^2$ Stuttgart Laboratory 1, Sony Semiconductor Solutions Europe, Sony Europe B.V.\\\
}

\twocolumn[{%
\renewcommand\twocolumn[1][]{#1}%
\maketitle
\begin{center}
	\centering
	\captionsetup{type=figure}
	\vspace{-20pt}
	\includegraphics[width=\textwidth]{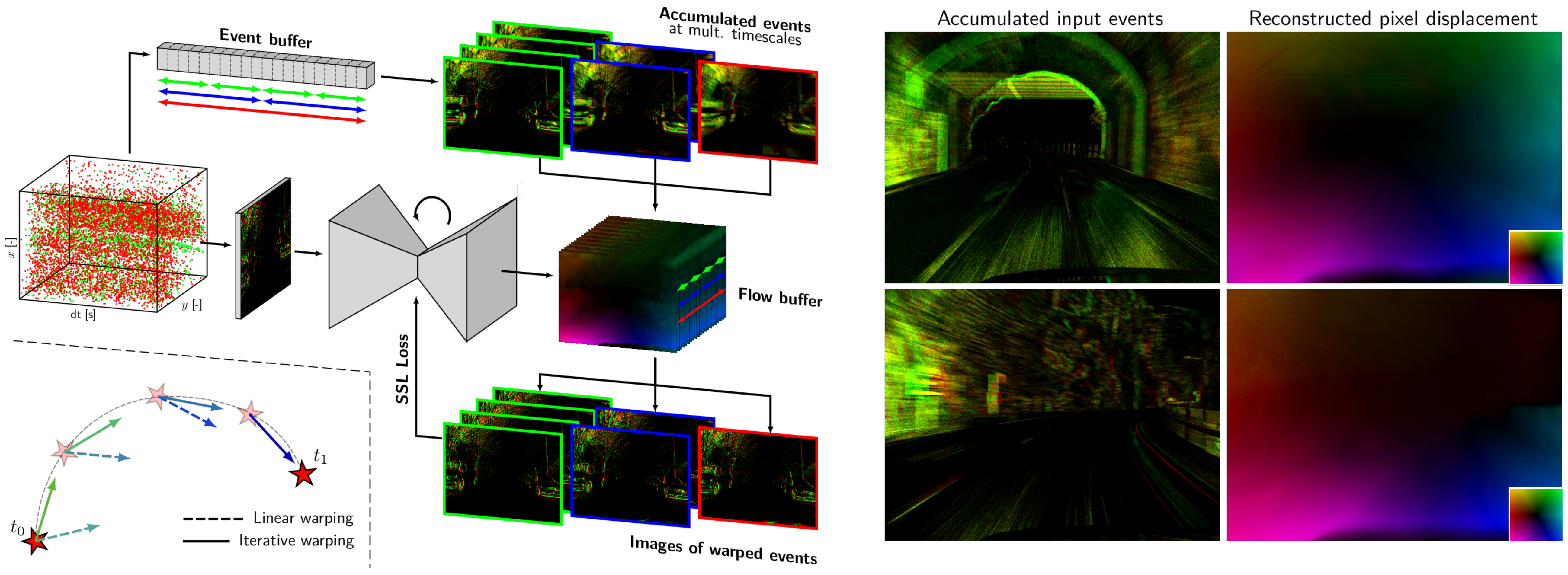}
	\captionof{figure}{Our pipeline estimates event-based optical flow by sequentially processing small partitions of the event stream with a recurrent model. We propose a novel self-supervised learning framework (left) based on a multi-timescale contrast maximization formulation that is able to exploit the high temporal resolution of event cameras via iterative warping to produce accurate optical flow predictions (right).}
	\label{fig:intro}
\end{center}%
}]

\maketitle
\ificcvfinal\thispagestyle{empty}\fi

\begin{abstract}
Event cameras have recently gained significant traction since they open up new avenues for low-latency and low-power solutions to complex computer vision problems. To unlock these solutions, it is necessary to develop algorithms that can leverage the unique nature of event data. However, the current state-of-the-art is still highly influenced by the frame-based literature, and usually fails to deliver on these promises. In this work, we take this into consideration and propose a novel self-supervised learning pipeline for the sequential estimation of event-based optical flow that allows for the scaling of the models to high inference frequencies. At its core, we have a continuously-running stateful neural model that is trained using a novel formulation of contrast maximization that makes it robust to nonlinearities and varying statistics in the input events. Results across multiple datasets confirm the effectiveness of our method, which establishes a new state of the art in terms of accuracy for approaches trained or optimized without ground truth.
\vfill
\end{abstract}

\vspace{-16pt}
\section{Introduction}
\label{sec:intro}
Event cameras capture per-pixel log-brightness changes at microsecond resolution \cite{gallego2019event}.\blfootnote{The project's code and additional material can be found at \url{https://mavlab.tudelft.nl/taming_event_flow/}.} This operating principle results in a sparse and asynchronous visual signal that, under constant illumination, directly encodes information about the apparent motion (i.e., optical flow) of contrast in the image space. These cameras offer several advantages, such as low latency and robustness to motion blur \cite{gallego2019event}, and hence hold the potential of a high-bandwidth 
estimation of this optical flow information. However, the event-based nature of the generated visual signal poses a paradigm shift in the processing pipeline, and traditional, frame-based algorithms become suboptimal and often incompatible. Despite this, the majority of learning-based methods that have been proposed so far for event-based optical flow estimation are still highly influenced by frame-based approaches. This influence is normally reflected in two key aspects of their pipelines: (i) the design of the network architecture, and (ii) the formulation of the loss function.

Regarding architecture design, most literature methods format subsets of the input events as dense volumetric representations \cite{zhu2019unsupervised} that are processed at once by stateless (i.e., non-recurrent) models \cite{zhu2019unsupervised, stoffregen2020train, paredes2021back, gehrig2021raft}. Similarly to their frame-based counterparts \cite{dosovitskiy2015flownet, sun2018pwc, teed2020raft}, these models estimate the per-pixel displacement over the time-length of the event volume using only the information contained within it. Consequently, these volumes need to encode enough spatiotemporal information for motion to be discernible. However, if done over relatively long time periods, the subsequent models suffer from limitations such as high latency or having to deal with large pixel displacements \cite{gehrig2021raft, gehrig2022dense,wu2022lightweight}.

With respect to the loss function, multiple options have been explored due to the lack of real-world datasets with per-event ground truth. Pure supervised learning can be used with datasets such as MVSEC \cite{zhu2018multivehicle} or DSEC-Flow \cite{gehrig2021raft}, but their ground truth only contains per-pixel displacement at low frequencies, which makes models difficult to train. On the other hand, a self-supervised learning (SSL) framework can be formulated using either the accompanying frames with the photometric error as a loss \cite{zhu2018ev, ding2022spatio, wan2022learning}, or through an events-only contrast maximization \cite{gallego2018unifying, gallego2019focus} proxy loss for motion compensation (i.e., event deblurring) \cite{mitrokhin2018event, zhu2019unsupervised, paredes2021back, hagenaars2021self, shiba2022secrets}. However, despite not relying on ground truth, all the literature on SSL for optical flow assumes that the events move linearly within the time window of the loss, which ignores much of the potential of event cameras and their high temporal resolution (see \figref{fig:intro}, bottom left).

In this work, we focus on the estimation of high frequency event-based optical flow and how this can be learned in an SSL fashion using contrast maximization with a relaxed linear motion assumption. To achieve this, we build upon the continuous-operation pipeline from Hagenaars \textit{et al}.\ \cite{hagenaars2021self}, which retrieves optical flow by \textit{sequentially} processing small partitions of the event stream with a stateful (i.e., recurrent) model over time, instead of dealing with large volumes of input events. We augment (and train) this pipeline with a novel contrast maximization formulation that performs event motion compensation in an iterative manner at multiple temporal scales, as shown in \figref{fig:intro}. Using this framework, we achieve the best accuracy of all contrast-maximization-based approaches on multiple datasets, only being outperformed by pure supervised learning methods trained with ground-truth data.

In summary, the extensions that we propose to the SSL framework in \cite{hagenaars2021self}, i.e., our main contributions, are:
\begin{itemize}
	\item The first iterative event warping module in the context of contrast maximization (see \secref{sec:nlew}). This module unlocks a novel multi-reference loss function that better captures the trajectory of scene points over time, thus improving the accuracy of the predictions.
    \item The first multi-timescale approach to contrast maximization, which adds robustness, improves convergence, and reduces tuning requirements of loss-related hyperparameters (see \secref{sec:timescale}).
\end{itemize}
As a result, we present the first self-supervised optical flow method for event cameras that relaxes the linear motion assumption, and hence that has the potential of exploiting the high temporal resolution of the sensor by producing estimates in a close-to continuous manner. We validate the proposed framework through extensive evaluations on multiple datasets. Additionally, we conduct ablation studies to show the effectiveness of each individual component.

\section{Related work}
\label{sec:related} 

Due to the aforementioned advantages of event cameras for optical flow estimation, extensive research has been carried out since these sensors were first introduced \cite{liu2018adaptive, benosman2012asynchronous, orchard2013spiking, benosman2013event, brosch2015event, bardow2016simultaneous, akolkar2020real, nagata2021optical, almatrafi2020distance, brebion2021real, shiba2022secrets}. Regarding learning-based approaches, the first method was proposed by Zhu \textit{et al}.\ in \cite{zhu2018ev} with EV-FlowNet: a UNet-like \cite{ronneberger2015u} architecture trained with SSL, with the supervisory signal coming from the photometric error between subsequent frames captured with an accompanying camera. To avoid the need for a secondary vision sensor, in \cite{zhu2019unsupervised}, Zhu \textit{et al}.\ proposed an SSL framework around the contrast maximization for motion compensation idea from \cite{gallego2018unifying, gallego2019focus}, hence relying solely on event data. This pipeline was used and further improved in \cite{paredes2021back, hagenaars2021self, shiba2022secrets}. Other approaches trained EV-FlowNet in a supervised fashion with synthetic/real ground-truth data and showed higher accuracy levels through evaluations on public benchmarks \cite{stoffregen2020train, gehrig2021raft}. However, they also highlighted EV-FlowNet's inability to deal with large pixel displacements \cite{gehrig2021raft}. Because of this, inspired by the frame-based literature \cite{teed2020raft}, Gehrig \textit{et al}.\ proposed E-RAFT in \cite{gehrig2021raft}: the first architecture to introduce the use of correlation volumes in the event camera literature. This model, which was recently augmented with attention mechanisms \cite{wang2022hanet}, achieved state-of-the-art performance in multiple datasets.

A novel perspective to learning event-based optical flow is to move away from event accumulation over long timespans, and instead rely on continuously-running stateful models that integrate information over time. The first methods of this kind were proposed by Paredes-Vall\'es \textit{et al}.\ \cite{paredes2020unsupervised} and Hagenaars \textit{et al}.\ \cite{hagenaars2021self}, but this idea has recently gained interest in the event camera literature \cite{ding2022spatio, ponghiran2022event, wu2022lightweight}. The reason is
that leveraging memory through sequential processing can potentially lead to lightweight and low-latency solutions that are also robust to large pixel displacements without the need for correlation volumes \cite{wu2022lightweight}. However, despite their potential of being scaled to high inference frequencies,
all these solutions assume that events move linearly in the timespan of their loss function, and hence cannot capture the true trajectory of scene points over time.

When it comes to learning nonlinear pixel trajectories from event data, only the work of Gehrig \textit{et al}.\ in \cite{gehrig2022dense} is to be highlighted. However, their approach uses multiple event and correlation volumes to fit B\'ezier curves to the trajectory of scene points, resulting in a high-latency solution. In contrast, in this work, we extend the continuous-operation pipeline from Hagenaars \textit{et al}.\ \cite{hagenaars2021self} with an SSL framework in which discrete trajectories are regressed at high frequency by leveraging memory within the models. This approach allows, for the first time, to exploit the high temporal resolution of event cameras while capturing more accurately the trajectory of scene points over time thanks to the proposed iterative event warping mechanism.

Among non-learning-based approaches, the work of Shiba \textit{et al}.\ in \cite{shiba2022secrets} bears particular relevance due to certain similarities with our framework. Specifically, Shiba \textit{et al}.\ propose a tile-based method for event-based optical flow estimation that extends contrast maximization \cite{gallego2018unifying, gallego2019focus} by incorporating (i) multiple spatial scales, (ii) a multi-reference focus loss, and (iii) a ``time-aware'' optical flow formulation. Similarly, our learning-based approach also employs a multi-scale strategy and utilizes a multi-reference focus loss. However, instead of making assumptions about the events' motion over extended time periods, we propose to learn their potentially nonlinear trajectories at high frequency by leveraging the iterative event warping module.

\section{Method}
\label{sec:method}

The goal of this work is to learn to sequentially estimate optical flow at high frequencies from a continuous stream of events. In such a pipeline, if the inference frequency is sufficiently high, events need to be processed nearly as soon as they are triggered by the sensor, with no pre-processing in between. Because of this, the integration of temporal information needs to happen in the network itself. To accomplish this, we propose the framework in \figref{fig:intro}, in which a stateful model is trained using our novel formulation of contrast maximization for sequential processing. The components of this framework are described in the following sections.

\subsection{Input format and contrast maximization}\label{sec:input}

For an ideal camera, an event $\boldsymbol{e}_i=(\boldsymbol{x}_i,t_i,p_i)$ of polarity $p_i\in\{+,-\}$ is triggered at pixel $\boldsymbol{x}_i=(x_i,y_i)^T$ and time $t_i$ whenever the change in log-brightness since the last event at that pixel location reaches the contrast sensitivity threshold for that polarity \cite{gallego2019event}.

As in \cite{hagenaars2021self}, we use a two-channel event count image as input representation, which gets populated with consecutive, non-overlapping, fine discrete partitions of the event stream
$\smash{\boldsymbol{\varepsilon}^{\text{inp}}_k\doteq\{\boldsymbol{e}_i\}}$
(further referred to as \textit{input partitions}), each containing all the events in a time window of a certain duration, i.e., $t_i \in [t^{\text{begin}}_k, t^{\text{end}}_k]$. This representation does not contain temporal information by itself, and recurrent models are hence required for estimating of optical flow.

Regarding learning, we use the contrast maximization framework \cite{gallego2019focus} to train, in an SSL fashion, to continuously estimate dense (i.e., per-pixel) optical flow from the event stream. Assuming brightness constancy, accurate flow information is encoded in the spatiotemporal misalignments (i.e., blur) among the events triggered by the same portion of a moving edge. To retrieve it, one has to compensate for this motion by geometrically transforming the events using a motion model. As in \cite{zhu2019unsupervised, paredes2021back, hagenaars2021self, shiba2022secrets}, we transport each event to a reference time $t_{\text{ref}}$ through:
\begin{equation}\label{eqn:motionmodel}
	\boldsymbol{x}'_i=\boldsymbol{x}_i + (t_{\text{ref}} - t_i)\boldsymbol{
		u}(\boldsymbol{x}_i)
\end{equation}
where $\smash{\boldsymbol{u}(\boldsymbol{x})=(u(\boldsymbol{x}),v(\boldsymbol{x}))^T}$ denotes the optical flow map used to transport each event from $t_i$ to $t_{\text{ref}}$. The result of aggregating the transformed events is further referred to as the image of warped events (IWE) at $t_{\text{ref}}$.

As the loss function of our SSL framework, we adopt the time-based focus objective function from \cite{hagenaars2021self}. Using the warped events at a given $t_{\text{ref}}$, we generate an image of the per-pixel average timestamps for each polarity $p'$ via bilinear interpolation:
\begin{align}\label{eqn:timeimage}
	\begin{aligned}
		T_{p'}(\boldsymbol{x}{;}\boldsymbol{u} |t_{\text{ref}}) &= \frac{\sum_{j} \kappa(x - x'_{j})\kappa(y - y'_{j})\bar{t}_{j}(t_{\text{ref}}, t_j)}{\sum_{j} \kappa(x - x'_{j})\kappa(y - y'_{j})+\epsilon}\\\kappa(a) &= \max(0, 1-|a|)\\
		j = \{i \mid p_{i}=&\ p'\}, \hspace{15pt}p'\in\{+,-\}, \hspace{15pt} \epsilon\approx 0
	\end{aligned}
\end{align}
where $\bar{t}_{i}$ denotes the normalized timestamp contribution of the $i$th event, according to \figref{fig:ts_norm} and \eqnref{eq:norm}.

Then, the contrast maximization loss function at $t_{\text{ref}}$ is defined as the scaled sum of the squared temporal images:
\begin{equation}\label{eq:scaling}
	\mathcal{L}_{\text{CM}}(t_{\text{ref}}) = \frac{\sum_{\boldsymbol{x}} T_{+}(\boldsymbol{x}{;}\boldsymbol{u} |t_{\text{ref}})^2 + T_{-}(\boldsymbol{x}{;}\boldsymbol{u} |t_{\text{ref}})^2}{\sum_{\boldsymbol{x}}\left[n(\boldsymbol{x}') > 0\right] + \epsilon}
\end{equation}
where $n(\boldsymbol{x}')$ denotes a per-pixel event count of the IWE. The lower the $\mathcal{L}_{\text{CM}}$, the better the event deblurring at $t_{\text{ref}}$.

As discussed in \cite{gallego2019focus, stoffregen2019event, hagenaars2021self}, for any focus objective function to be a robust supervisory signal for contrast maximization, the event partition used in the optimization needs to contain enough motion information (i.e., blur) so it can be compensated for. However, as in \cite{hagenaars2021self}, that is not the case in our input partitions due to the fine discretization of the event stream that we are targeting. Therefore, only at training time, we define the so-called \textit{training partition} (or event buffer in \figref{fig:intro}) $\smash{\boldsymbol{\varepsilon}^{\text{train}}_{k\rightarrow k+R}\doteq\{\boldsymbol{\varepsilon}^{\text{inp}}_{i}\}_{i=k}^{k+R}}$, which stacks together the events in the input partitions of $R$ successive forward passes through the network. Once these are performed, we use this partition and the estimated optical flow maps to compute the loss,
and use truncated backpropagation through time to update the model parameters. After this update, we detach the states of the network from the computational graph and clear the training partition and optical flow buffer. The importance of sequential processing, as an alternative to training stateless models in short input partitions, is corroborated in the supplementary material.

\begin{figure}[t]
	\centering
	\includegraphics[width=0.451\textwidth]{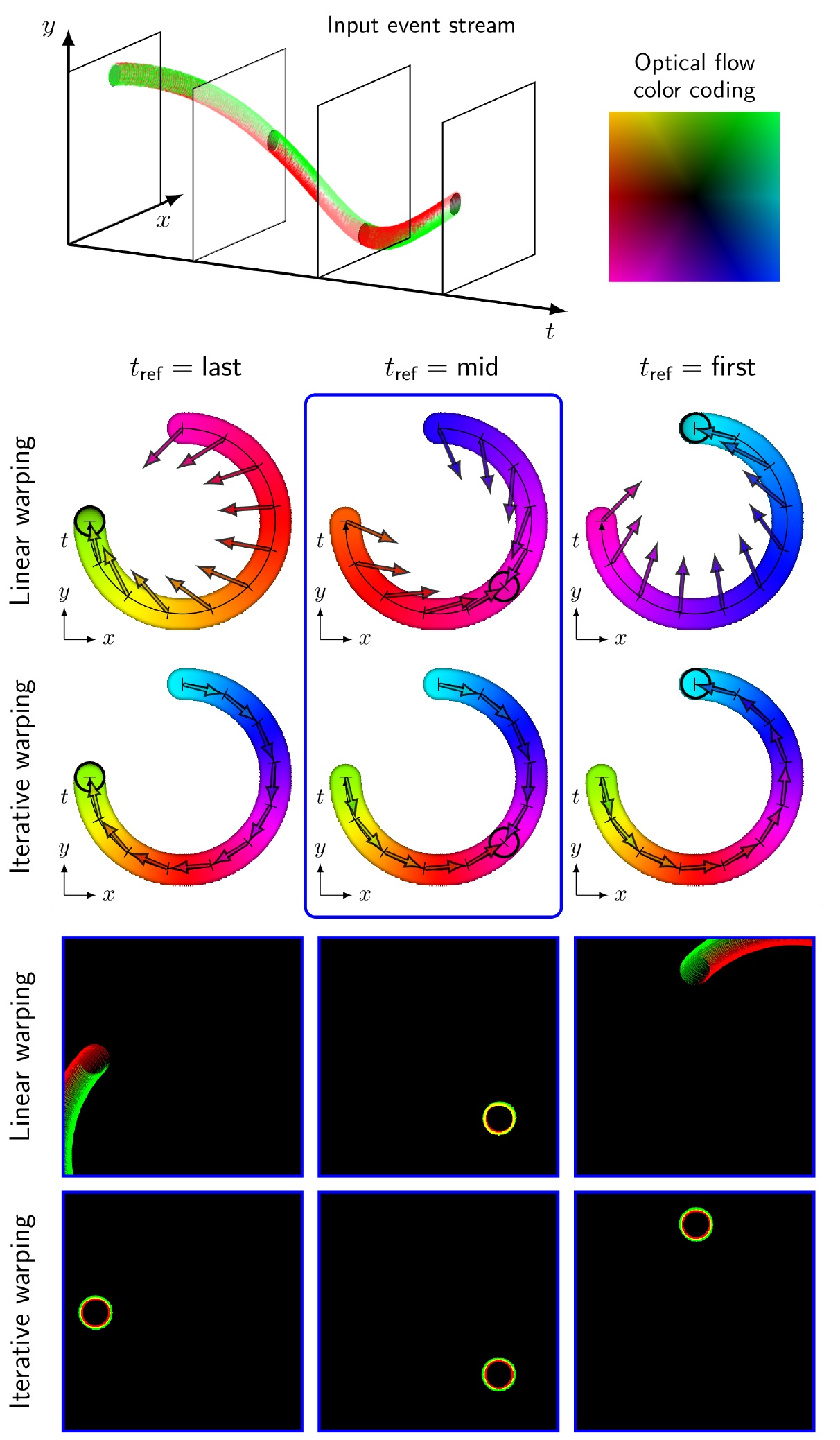}
	\caption{Incompatibility of multi-reference deblurring and linear event warping in the presence of nonlinearities in the pixel trajectories. \textit{Top}: Events generated by a moving dot following circular motion (left), and optical flow color-coding scheme (right). \textit{Middle}: Optical flow solutions required to produce sharp IWEs at different $t_{\text{ref}}$ using linear and iterative warping. While the former requires a different solution for each $t_{\text{ref}}$, the proposed iterative warping can achieve this using a single solution. Arrows illustrate the direction of the required displacement $\Delta\boldsymbol{x}_i = (t_{\text{ref}} - t_i)\boldsymbol{u}(\boldsymbol{x}_i)$ at each (discretized) spatial location. \textit{Bottom}: Resulting IWEs at different $t_{\text{ref}}$ using the optimal optical flow map for $t_{\text{ref}}=\text{mid}$.
	}
	\label{fig:lin_nonlin}
\end{figure}

\subsection{Iterative event warping}\label{sec:nlew}

In order to better approximate the trajectories of scene points over time, in this work we relax the linear motion assumption in the SSL framework for event-based optical flow by augmenting the sequential-estimation pipeline from \cite{hagenaars2021self} with \textit{iterative event warping}. Instead of transporting events to a given $t_{\text{ref}}$ assuming linear motion
regardless of the length of the warping interval (as in \cite{zhu2019unsupervised, paredes2021back, hagenaars2021self, shiba2022secrets}), we perform a finer discretization of the event trajectories and assume that motion is only linear between optical flow estimates. Therefore, to express a group of events at a given $t_{\text{ref}}$, we geometrically transform them using all the intermediate optical flow estimates through multiple iterations of \eqnref{eqn:motionmodel}.
\figref{fig:lin_nonlin} shows the differences between the linear event warping in \cite{hagenaars2021self} and the proposed iterative augmentation, as well as the limitations of the former. Note that our iterative warping is fundamentally different from that of the work of Wu \textit{et al}.\ in \cite{wu2022lightweight}, where input events are deblurred before being passed to the models using residual optical flow estimates until convergence.

\begin{figure}[!t]
	\centering
	\includegraphics[width=0.48\textwidth]{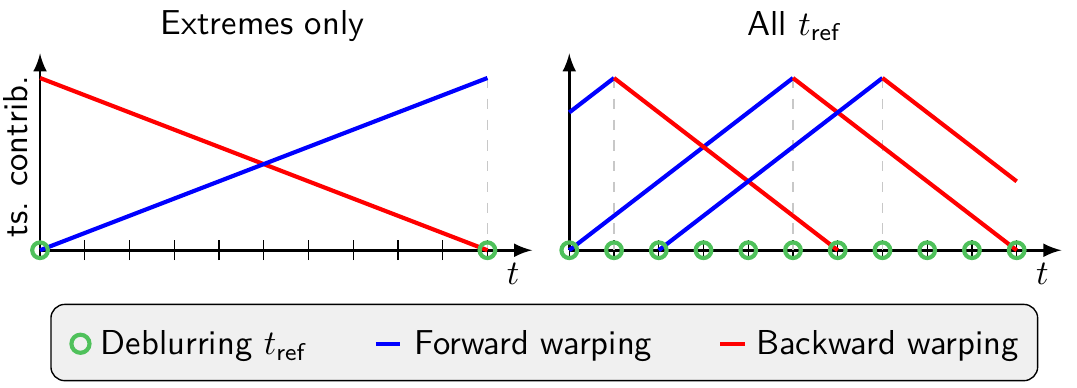}
	\caption{Timestamp normalization profiles for the per-event contributions to the images of average timestamps in \eqnref{eqn:timeimage}, for $R$$=$$10$. \textit{Left}: Deblurring only done at the extremes of the training partition, as in \cite{zhu2019unsupervised, paredes2021back, hagenaars2021self}. \textit{Right}: Deblurring done at all reference times, thanks to the proposed iterative event warping. Normalization profiles only shown for three $t_{\text{ref}}$ for a better visualization.}
	\label{fig:ts_norm}
\end{figure}

Literature methods on event-based optical flow with contrast maximization compute the focus objective function at multiple reference times in the training partition (usually at the extremes) to prevent overfitting and/or scaling issues during backpropagation \cite{zhu2019unsupervised, paredes2021back, hagenaars2021self, shiba2022secrets}. However, since these approaches assume optical flow constancy in the span of their loss function, they suffer with nonlinearities in the pixel trajectories (see the blurry IWEs in \figref{fig:lin_nonlin}, bottom). On the contrary, because of the finer discretization of the event trajectories coming with our sequential processing pipeline and the proposed iterative warping, we can use any (combination of) reference time(s) for the computation of the focus objective function. In fact, as shown in \figref{fig:ts_norm} (right), we propose the use of \textit{all} the discretization points as reference times for event deblurring. Apart from the aforementioned regularizing benefits, having to produce sharp IWEs at any $t_{\text{ref}}$ forces the models to estimate a sequence of optical flow maps that is consistent with the velocity profile of the event stream. For a given training partition of length $R$, the loss is computed as follows:
\begin{equation}\label{eq:tref}
	\mathcal{L}_{\text{CM}}^R = \frac{1}{R+1}\sum_{t_{\text{ref}}=0}^{R}\mathcal{L}_{\text{CM}}(t_{\text{ref}})
\end{equation}
which ensures that the IWEs (and the corresponding images of average timestamps in \eqnref{eqn:timeimage}) at all reference times $t_{\text{ref}}\in [0, R]$ contribute equally to the loss.

As in \cite{zhu2019unsupervised, paredes2021back, hagenaars2021self}, for $\mathcal{L}_{\text{CM}}(t_{\text{ref}})$ in \eqnref{eq:scaling} to be a valid supervisory signal, events that are temporally close to the reference time $t_{\text{ref}}$ need to contribute to the temporal image in \eqnref{eqn:timeimage} with larger timestamp values than events that are far in time. Therefore, once the events have been transported to $t_{\text{ref}}$, their timestamp is normalized prior to the computation of \eqnref{eqn:timeimage} as follows:
\begin{equation}\label{eq:norm}
	\bar{t}_i (t_{\text{ref}}, t_i) = 1 - \frac{|t_{\text{ref}} - t_i|}{R}, \quad \text{with } t_i \in [0, R]
\end{equation}
which results in the normalization profiles in \figref{fig:ts_norm}, for a given training partition of length $R$.

\begin{figure}[!t]
	\centering
	\includegraphics[width=0.5\textwidth]{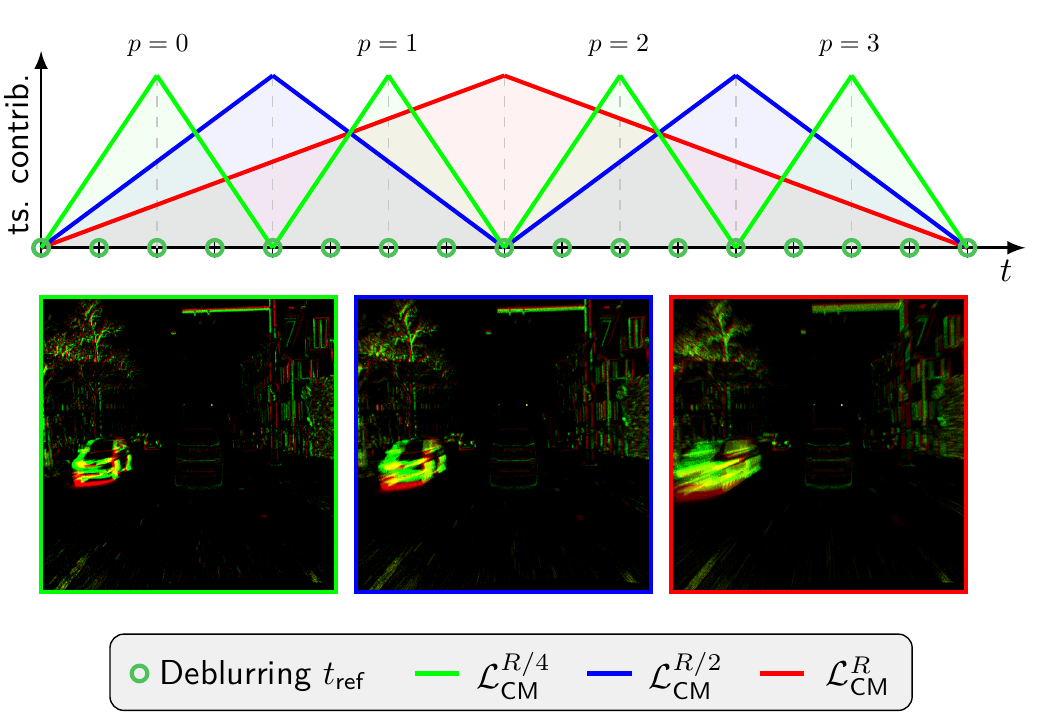}
	\caption{Multi-timescale approach to contrast maximization. For a given training partition of length $R$, we fit multiple sub-partitions of different lengths (in the figure: one of length $R$ in red, two $R/2$ in blue, and four $R/4$ in green) and compute the loss in each of them according to \eqnref{eq:tref}. The global loss is computed as in \eqnref{eq:multits}. This figure only shows the timestamp normalization profiles of the central $t_{\text{ref}}$ of each sub-partition, but the losses are still computed at all reference times. An image representation of the accumulated input events in a sub-partition of each timescale is also shown.
	}
	\label{fig:ts_multiple}
\end{figure}

Inspired by \cite{meister2018unflow}, we mask individual events from the computation of the loss 
whenever we detect that they are transported outside the image space through the event warping process in the span of a deblurring window defined at the reference time $t_{\text{ref}}$. This prevents our models from learning incorrect deformations at the image borders, and is motivated by the fact that the complete trajectory of the pixel is only partially observable in the training partition. 
An ablation study on the impact of this masking strategy can be found in the supplementary material.

Lastly, we do not augment the loss function in \eqnref{eq:tref} with smoothing priors acting as regularization mechanisms. With iterative event warping, the error propagates through all the pixels covered in the warping process, regardless of whether they have input events or not. Therefore, the spatial coherence of the resulting optical flow maps is enhanced.

\subsection{Deblurring at multiple timescales}\label{sec:timescale}

Despite the addition of iterative event warping, 
the success of our SSL framework still heavily depends on the hyperparameters that control the amount of motion information perceived by the networks in the span of a deblurring window. In our pipeline, these are: $\text{dt}_{\text{input}}$, the timestep used to discretize the event stream; and $R$, the number of forward passes, and hence optical flow maps, per loss. Thus, the effective length (in units of time) of the event window used for motion compensation is given by $\text{dt}_{\text{input}}\times R$. We hypothesize that, for each training dataset, there is an optimal length for this window that depends on the statistics of the data (e.g., event density, distribution of optical flow magnitudes) and model architecture, and that deviations from this optimal length lead to the learning of suboptimal solutions. E.g., shorter windows may converge to solutions that are more selective to fast rather than slow moving objects, 
and vice versa. Note that not only our method is sensitive to the tuning of these parameters, but also previous approaches based on contrast maximization \cite{zhu2019unsupervised, paredes2021back, hagenaars2021self, shiba2022secrets}.

To add a layer of robustness to the framework and relax its strong dependency on hyperparameter optimization, we propose the \textit{multi-timescale approach} illustrated in \figreftwo{fig:intro}{fig:ts_multiple}. For a given training partition of length $R$, instead of computing a single focus loss through \eqnref{eq:tref}, we compute this loss at $S$ temporal scales of length $R/2^s$, with $0\leq s \leq S-1$, and combine them as follows:
\begin{equation}\label{eq:multits}
	\mathcal{L}_{\text{CM}}^\text{multi} = \frac{1}{S}\sum_{s=0}^{S-1}\frac{1}{2^s}\sum_{p=0}^{2^s-1}\mathcal{L}_{\text{CM}, p}^{R/2^{s}}
\end{equation}
As shown, we fit multiple non-overlapping sub-partitions in the training buffer if $s>0$. The subscript $p$ indicates their location in this buffer, starting from the earliest (see \figref{fig:ts_multiple}).

Note that, through this multi-timescale approach to contrast maximization, our models need to converge to a solution that is suitable for all the timescales in the optimization, regardless of their length. An alternative formulation would be to incorporate per-pixel learnable masks (i.e., an attention module in the loss space) so that, depending on the input statistics, learning only happens at the most adequate scale. However, for this to happen, the loss function would have to be augmented to stimulate this behavior, and it is unclear how that would be done in practice.

\subsection{Network architecture}\label{sec:arch}

\begin{figure}[t]
	\centering
	\includegraphics[width=0.45\textwidth]{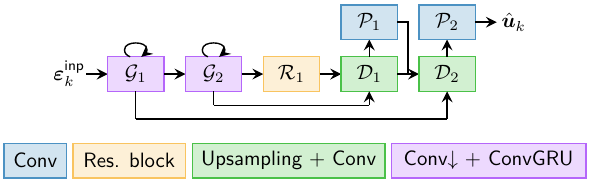}
	\caption{Schematic of the model architecture used in this work. It is characterized by $N_{\mathcal{G}}$ recurrent encoders, $N_{\mathcal{R}}$ residual blocks, and $N_{\mathcal{G}}$ decoder layers. Optical flow estimates are produced at all decoder levels. In this diagram, $N_{\mathcal{G}}=2$ and $N_{\mathcal{R}}=1$.}
	\label{fig:archs}
\end{figure}

We use the recurrent version of EV-FlowNet \cite{zhu2018ev} proposed in \cite{hagenaars2021self}\footnote{Our architecture is equivalent to ConvGRU-EV-FlowNet \cite{hagenaars2021self}. However, for the purpose of clarity within the rest of the paper, we will use this term to specifically refer to the original model from \cite{hagenaars2021self}.} (see \figref{fig:archs}). The events are represented as event count images (see \secref{sec:input}), then passed through four encoders with strided convolutions followed by ConvGRUs \cite{ballas2015delving} (channels doubling, starting from 64), two residual blocks \cite{he2016deep}, and then four decoders performing bilinear upsampling followed by convolution. After each decoder, there is a skip connection (using element-wise summation) from the corresponding encoder, as well as a depthwise convolution to produce estimates at lower scales, which are then concatenated with the activations of the previous decoder. Note that the proposed focus loss function (see \eqnref{eq:multits}) is applied to each intermediate optical flow estimate via upsampling. Lastly, all layers use $3\times 3$ kernels and ReLU activations except for the prediction layers, which use TanH. 

\section{Experiments}
\label{sec:experiments}

We evaluate our method on the DSEC-Flow \cite{gehrig2021dsec, gehrig2021raft} and MVSEC \cite{zhu2018multivehicle} datasets. We evaluate the accuracy of the predictions based on the following metrics: (i) EPE (lower is better, $\downarrow$), the endpoint error; (ii) $\%_{3\text{PE}}$ ($\downarrow$), the percentage of points with EPE greater than 3 pixels; (iii) FWL ($\uparrow$) \cite{stoffregen2020train}, a deblurring quality metric based on the variance of the IWEs; and (iv) RSAT ($\downarrow$) \cite{hagenaars2021self}, a deblurring quality metric based on the per-pixel average timestamps of the IWEs. We compare our solution to the published baselines, which range from supervised learning (SL) methods trained with ground truth, to SSL methods trained with grayscale images (SSL$_{\text{F}}$) or events (SSL$_{\text{E}}$), and model-based approaches (MB).

We train all our models on a subset of sequences from the training dataset of DSEC-Flow (only daylight recordings, see supplementary material). This corresponds to 19 minutes of training data, which we split into 572 $128\times128$ (randomly-cropped) sequences of 2 seconds each. We use a batch size of 8 and train until convergence with the Adam optimizer \cite{kingma2014adam} and a learning rate of $1\mathrm{e}{}$-$5$. To keep memory usage within limits, we only propagate error gradients through up to $1\mathrm{e}{}3$ randomly-chosen events per millisecond of data. Despite this, note that we warp and use all the input events for the computation of the loss function.

\subsection{Evaluation procedure}\label{sec:recon}

\begin{figure}[t]
	\centering
	\includegraphics[width=0.485\textwidth]{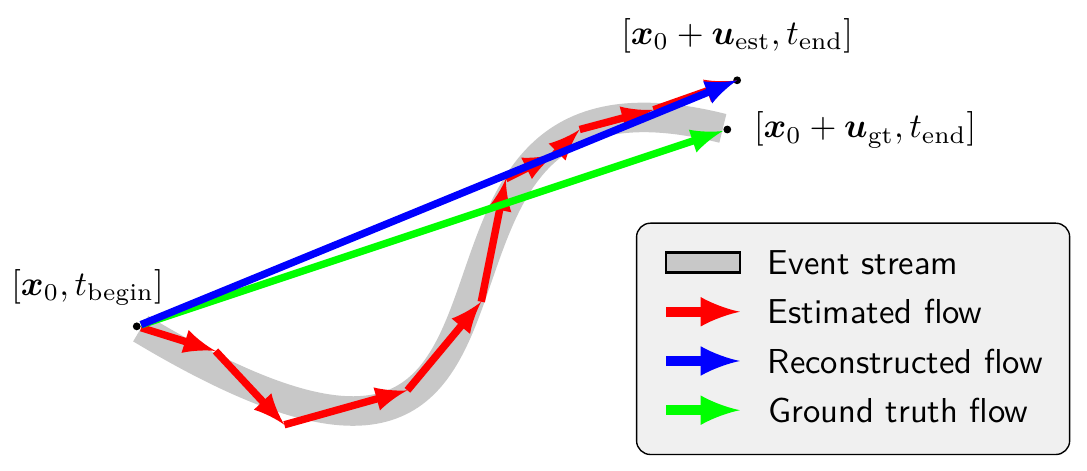}
	\caption{Reconstruction of the pixel displacement of a scene point in the time window of a ground-truth sample from the multiple optical flow maps estimated in this period, i.e., $\smash{\forall \boldsymbol{u}_k\in[t_{\text{begin}}, t_{\text{end}}]}$. The error of the last optical flow estimate is magnified for clarity.
	}
	\label{fig:flow_recon}
\end{figure}

\begin{table}[!t]
	\centering
	\resizebox{1\linewidth}{!}{%
	{\renewcommand{\arraystretch}{1.1} 
	\begin{tabular}{lcccc}
		\thickhline
		\thickhline
		& EPE$\downarrow$ & $\%_{3\text{PE}}$$\downarrow$& FWL$\uparrow$& RSAT$\downarrow$\\\thickhline
		\parbox[t]{0mm}{\multirow{8}{*}{\rotatebox[origin=c]{90}{\footnotesize SL}}}
		\hspace{10pt}E-RAFT \cite{gehrig2021raft}& 0.79 & 2.68 & 1.33 & \underline{0.87} \\
		\hspace{12.5pt}EV-FlowNet, Gehrig \textit{et al}.\ \cite{gehrig2021raft}& 2.32 & 18.60 & -&- \\
        \hspace{12.5pt}Gehrig \textit{et al}.\ \cite{gehrig2022dense}& 0.75 & 2.44 & -&- \\
		\hspace{12.5pt}IDNet \cite{wu2022lightweight}& \textbf{0.72} & \textbf{2.04} & -&- \\
		\hspace{12.5pt}TIDNet \cite{wu2022lightweight}& 0.84 & 3.41 & -&- \\
        \hspace{12.5pt}TMA \cite{liu2023tma}& \underline{0.74} & \underline{2.30} & -&- \\
		\hspace{12.5pt}Cuadrado \textit{et al}.\ \cite{cuadrado2023optical}& 1.71 & 10.31 & -&- \\
		\hspace{12.5pt}E-Flowformer \cite{li2023blinkflow}& 0.76 & 2.68 & -&- \\
		\hdashline
		\parbox[t]{0mm}{\multirow{8}{*}{\rotatebox[origin=c]{90}{\footnotesize SSL$_{\text{E}}$}}}
		\hspace{10pt}EV-FlowNet$^{*}$ \cite{zhu2019unsupervised}  & 3.86 & 31.45 & 1.30 & \textbf{0.85}\\
		\hspace{12.5pt}ConvGRU-EV-FlowNet$^{*}$ \cite{hagenaars2021self} & 4.27 & 33.27 & \underline{1.55} & 0.90\\
		\hspace{12.5pt}dt $=0.01$s, $R = 2$, $S=1$ (Ours) & 9.66 & 86.44 & \textbf{1.91} & 1.07\\
		\hspace{12.5pt}dt $=0.01$s, $R = 5$, $S=1$ (Ours) & 4.05 & 52.22 & 1.58 & 0.97\\
		\hspace{12.5pt}dt $=0.01$s, $R = 10$, $S=1$ (Ours) & 2.33 & 17.77 & 1.26 & 0.88\\
		\hspace{12.5pt}dt $=0.01$s, $R = 20$, $S=1$ (Ours) & 16.63 & 33.67 & 1.06 & 1.10\\
		\hspace{12.5pt}dt $=0.01$s, $R = 10$, $S=3$ (Ours) & 2.82 & 27.09 & 1.37 & 0.92\\
		\hspace{12.5pt}dt $=0.01$s, $R = 20$, $S=4$ (Ours) & 2.73 & 23.73 & 1.24 & 0.90\\
		\hdashline
		\parbox[t]{0mm}{\multirow{1.35}{*}{\rotatebox[origin=c]{90}{\footnotesize MB}}}
		\vspace{-10pt}\\
		\hspace{12.5pt}Shiba \textit{et al}.\ \cite{shiba2022secrets}& 3.47 & 30.86 & 1.37 & 0.89 \\
		\vspace{-12.5pt}\\
		\thickhline
		\thickhline
		\multicolumn{4}{l}{\hspace{12.5pt}\small $^*$Retrained by us on DSEC-Flow, linear warping.}
	\end{tabular}}}
	\caption{Quantitative evaluation on the DSEC-Flow dataset \cite{gehrig2021raft}. Best in bold, runner up underlined. A breakdown of the results is provided in the supplementary material. 
 }
	\label{tab:DSEC}
	\vspace{-5pt}
\end{table}

\begin{figure*}[t]
	\centering
	\includegraphics[width=0.95\textwidth]{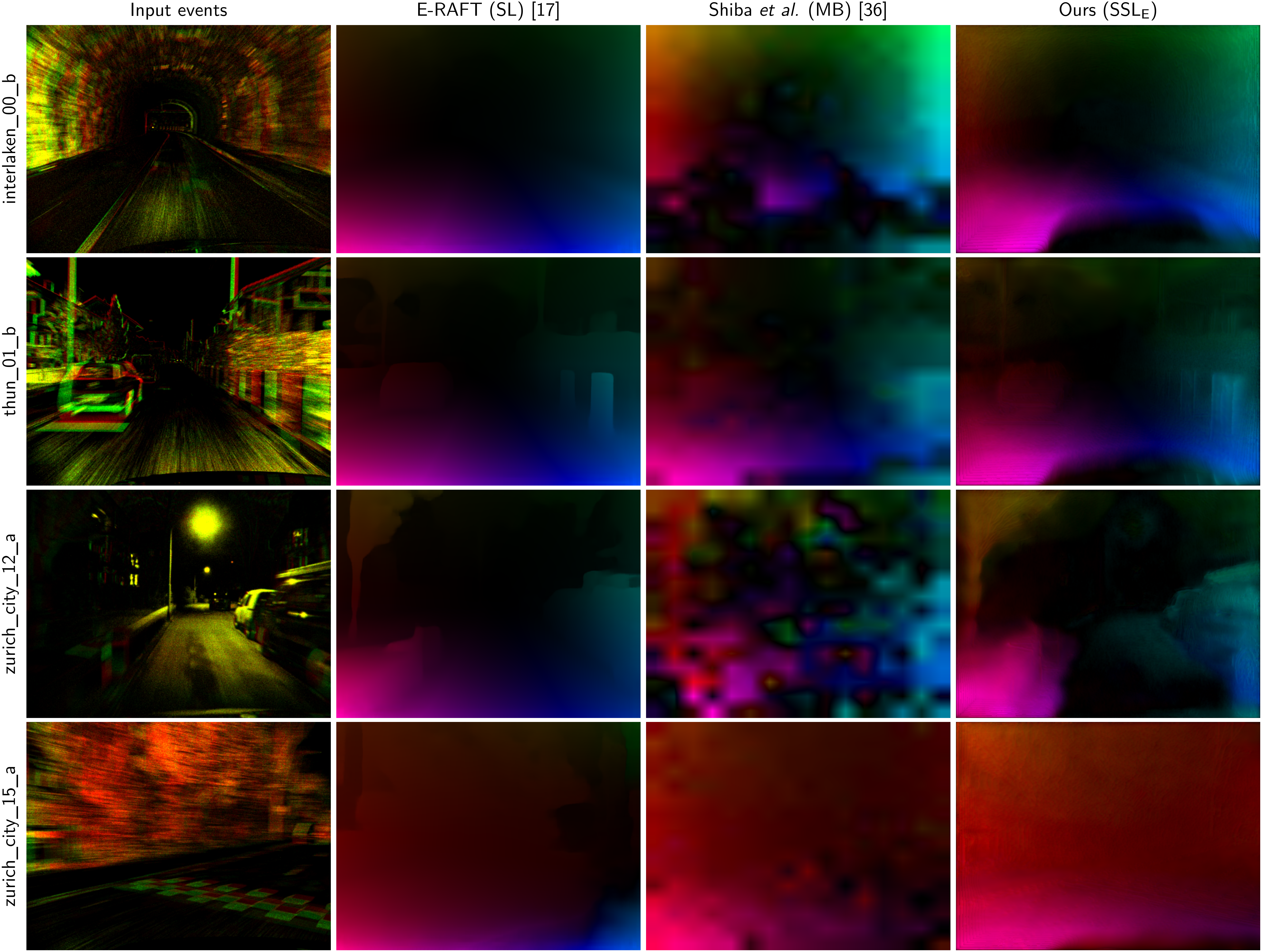}
	\caption{Qualitative comparison of our method with the state-of-the-art E-RAFT architecture \cite{gehrig2021raft} and the model-based approach from Shiba \textit{et al}.\ \cite{shiba2022secrets} on sequences from the test partition of the DSEC-Flow dataset \cite{gehrig2021raft}. Ground truth not included due to unavailability. The optical flow color coding can be found in \figref{fig:lin_nonlin} (top), and the corresponding IWEs in the supplementary material.}
	\label{fig:DSEC}
\end{figure*}

When evaluating our sequential models, if $\text{dt}_{\text{gt}}>\text{dt}_{\text{input}}$, we need to reconstruct the estimated per-pixel displacement in the ground-truth time window from the multiple optical flow maps estimated in this period. We do this by first averaging the (bilinearly interpolated) optical flow vectors that describe the trajectory of each scene point, and then by scaling the resulting optical flow vectors by $\text{dt}_{\text{gt}}/\text{dt}_{\text{input}}$. This converts them from units of pixels/input to units of pixel displacement. An illustration of this reconstruction is shown in \figref{fig:flow_recon} for a scene point following a nonlinear trajectory. Note that our solution is subject to cumulative errors when evaluated through this reconstruction on benchmarks with ground truth provided at low rates (e.g., 10 Hz in DSEC-Flow \cite{gehrig2021raft}). Therefore, it will compare unfavorably to other, non-sequential, methods that only produce a single optical flow estimate in the timespan of a ground-truth sample.

\subsection{Optical flow evaluation}\label{sec:evaluation}

\textbf{Evaluation on DSEC-Flow:} Quantitative results of our evaluation on DSEC-Flow are presented in \tabref{tab:DSEC}, and are supported by the qualitative comparison in \figref{fig:DSEC}. For this experiment, we trained multiple models with the same dt$_{\text{input}}=$ 0.01s (i.e., $\times$10 faster than DSEC's ground truth) but different lengths of the training partition, and with and without the multi-timescale approach. Multiple conclusions can be derived from the reported results. Firstly, our best performing model (i.e., $R$ $=$ $10$, $S$ $=$ $1$) achieves the best accuracy of all contrast-maximization-based approaches on this dataset according to the EPE and the percentage of outliers.
Specifically, it outperforms the baselines with an improvement in the EPE in the 33\%\hspace{2pt}--\hspace{2pt}45\% range, only being outperformed by SL methods trained with ground truth on the same dataset \cite{gehrig2021raft,wu2022lightweight,liu2023tma, cuadrado2023optical, li2023blinkflow}. This confirms that (i) the timestamp-based loss function in \secref{sec:input} allows us to learn accurate event-based optical flow (contrary to the findings of \cite{shiba2022fast}); and that (ii) our augmentations 
to the sequential pipeline in \cite{hagenaars2021self} lead to a significant improvement in the accuracy of the model (i.e., 45\% improvement in the EPE).

Secondly, these results also confirm our hypothesis that, for each training dataset, there is an optimal length for the training partition $R$ in terms of the EPE. According to \tabref{tab:DSEC}, the optimal $R$ for this dataset, our model architecture, and our dt$_{\text{input}}$ is 10 (i.e., 0.1s of event data), with the EPE increasing if the training partition is made shorter or longer. However, as also shown in this table, we can relax the strong dependency of the contrast maximization framework on this parameter through the proposed multi-timescale approach. Our $S$ $>$ 1 models converged to solutions that slightly underperform our best performing single-scale model (EPE went up by 17\%\hspace{2pt}--\hspace{2pt}21\%), but were trained without the need to fine-tune the length of the training partition. Note that, despite this slight drop in accuracy, these multi-timescale solutions still outperform the other non-SL baselines. To further support these results, a visualization of the distribution of the EPE of our models as a function of the ground truth magnitude is provided in the supplementary material.

Lastly, \tabref{tab:DSEC} also allow us to conclude that deblurring quality metrics FWL \cite{stoffregen2020train} and RSAT \cite{hagenaars2021self} are not reliable indicators of the quality of the estimated optical flow. The reason for this is their inability to capture ``event collapse'' issues (as described in \cite{shiba2022event}), and would give favorable scores to undesirable solutions that warp all the events into a few pixels. According to our results, the FWL metric, being the spatial variance of the IWE relative to that of the identity warp, suffers more from this issue: the best FWL value is obtained with a model with 9.66 EPE.

\begin{table}[!t]
	\vspace{-2.5pt}
	\centering
	\resizebox{0.75\linewidth}{!}{%
	{\renewcommand{\arraystretch}{1.1} 
	\begin{tabular}{lcccc}
		\thickhline
		\thickhline
		& EPE$\downarrow$ & $\%_{3\text{PE}}$$\downarrow$& FWL$\uparrow$& RSAT$\downarrow$\\\thickhline
		dt $=0.1$s$^{*}$ & 3.48 & 34.72 & 0.98 & \textbf{0.87} \\
		dt $=0.05$s & 3.09 & 27.36 & 1.11 & 0.91 \\
		dt $=0.01$s & \textbf{2.33} & \textbf{17.77} & 1.26 & \underline{0.88} \\
		dt $=0.005$s & \underline{2.34} & \underline{17.92} & \underline{1.38} & 0.89 \\
		dt $=0.002$s & 2.66 & 21.83 & \textbf{2.04} & 0.90 \\
		\thickhline
		\thickhline
		\multicolumn{5}{l}{\hspace{0pt}\small $^*$Non-recurrent, volum.\ event repr.\ with 10 bins.}
	\end{tabular}}}
	\caption{Impact of the input window length on the DSEC-Flow dataset \cite{gehrig2021raft}. Best in bold, runner up underlined.}
	\label{tab:time}
	\vspace{-5pt}
\end{table}

Regarding qualitative results, \figref{fig:DSEC} shows a comparison of our best performing model with the state-of-the-art E-RAFT architecture \cite{gehrig2021raft} and the contrast-maximization-based approach from Shiba \textit{et al}.\ \cite{shiba2022secrets} on multiple sequences from the test partition of DSEC-Flow (i.e., ground truth is unavailable). These results confirm that our models are able to estimate high quality event-based optical flow despite not having access to ground-truth data during training, and also show the superiority of our method over the current best contrast-maximization-based approach \cite{shiba2022secrets}. Two limit cases in which our models provide suboptimal solutions are also shown in this figure: (i) sequences recorded at night (e.g., zurich\_city\_12\_a) due to the presence of large amounts of events triggered by flashing lights and not by motion; and (ii) the car hood, which is also problematic for E-RAFT (i.e., does not capture it) and for \cite{shiba2022secrets}. 
Note that, in our case, (ii) is an artifact of the pixel displacements reconstructed from multiple optical flow estimates, and could be mitigated by having an occlusion handling mechanism in this reconstruction process.

In addition to the evaluation in \tabref{tab:DSEC} and \figref{fig:DSEC}, we also conducted an experiment in which we trained multiple models with different dt$_{\text{input}}$ (ranging from 0.1s to 0.002s) but with the same amount of information in the training partition: 0.1s of event data. Results in \tabref{tab:time} show that our sequential pipeline
leads to an improvement in the accuracy of the predicted optical flow maps with respect to the stateless EV-FlowNet, which processes the 0.1s of event data at once. This improvement is due to the fact that the complexity of dealing with large pixel displacements gets reduced when processing the input data sequentially using shorter input windows. In addition to this, \tabref{tab:time} also shows that the accuracy of our models is not compromised when estimating optical flow at higher frequencies, despite the high sparsity levels in the input data at those rates. 


\begin{table}[!t]
	\vspace{-5pt}
	\centering
	\resizebox{0.85\linewidth}{!}{%
	{\renewcommand{\arraystretch}{1.1} 
	\begin{tabular}{lcc}
		\thickhline
		\thickhline
		& EPE$\downarrow$& $\%_{\text{3PE}}$$\downarrow$\\\thickhline
		\parbox[t]{0mm}{\multirow{5}{*}{\rotatebox[origin=c]{90}{SL}}}
		\hspace{10pt}EV-FlowNet+ \cite{stoffregen2020train}& 0.68 & 0.99 \\
		\hspace{12.5pt}E-RAFT \cite{gehrig2021raft}& \textbf{0.24} & 1.70 \\
		\hspace{12.5pt}EV-FlowNet, Gehrig \textit{et al}.\ \cite{gehrig2021raft}& 0.31 & \textbf{0.00} \\
        \hspace{12.5pt}TMA \cite{liu2023tma}& \underline{0.25} & 0.07 \\
        \hspace{12.5pt}Cuadrado \textit{et al}.\ \cite{cuadrado2023optical}& 0.85 & -\\
		\hdashline
		\parbox[t]{0mm}{\multirow{2}{*}{\rotatebox[origin=c]{90}{SSL$_{\text{F}}$}}}
		\hspace{10pt}EV-FlowNet, Zhu \textit{et al}.\ \cite{zhu2018ev}& 0.49 & 0.20 \\
		\hspace{12.5pt}Ziluo \textit{et al}.\ \cite{ding2022spatio}& 0.42 & \textbf{0.00} \\
		\hdashline
		\parbox[t]{0mm}{\multirow{5}{*}{\rotatebox[origin=c]{90}{SSL$_{\text{E}}$}}}
		\hspace{10pt}EV-FlowNet, Zhu \textit{et al}.\ \cite{zhu2019unsupervised}& 0.32 & \textbf{0.00}  \\
		\hspace{12.5pt}EV-FlowNet, Paredes-Vall\'es \textit{et al}.\ \cite{paredes2021back}& 0.92 & 5.40  \\
		\hspace{12.5pt}EV-FlowNet, Shiba \textit{et al}.\ \cite{shiba2022secrets}& 0.36 & 0.09 \\
		\hspace{12.5pt}ConvGRU-EV-FlowNet \cite{hagenaars2021self}& 0.47 & 0.25  \\
		\hspace{12.5pt}Ours & 0.27 & \underline{0.05} \\  
		\hdashline
		\parbox[t]{0mm}{\multirow{3}{*}{\rotatebox[origin=c]{90}{MB}}}
		\hspace{10pt}Akolkar \textit{et al}.\ \cite{akolkar2020real}& 2.75 & - \\
		\hspace{12.5pt}Brebion \textit{et al}.\ \cite{brebion2021real}& 0.53 & 0.20 \\
		\hspace{12.5pt}Shiba \textit{et al}.\ \cite{shiba2022secrets}& 0.30 & 0.11 \\
		\thickhline
		\thickhline
	\end{tabular}}}
	\caption{Quantitative evaluation on MVSEC's outdoor\_1 sequence \cite{zhu2018multivehicle}. Best in bold, runner up underlined. Results on the indoor sequences can be found in the supplementary material. 
 }
 \label{tab:mvsec}
\end{table}

\textbf{Evaluation on MVSEC:} Quantitative results of our evaluation on the oudoor\_day1 sequence from MVSEC are presented in \tabref{tab:mvsec}, and are supported by the qualitative comparison in the supplementary material. For this experiment, since (i) there is no consensus in the literature with respect to the training dataset \cite{zhu2018ev, zhu2019unsupervised, stoffregen2020train, paredes2021back, hagenaars2021self, gehrig2021raft, shiba2022secrets}, and (ii) the outdoor\_day2 sequence (i.e., the other daylight, automotive sequence) is only 9 minutes of duration during which the event camera is subject to high frequency vibrations \cite{zhu2018multivehicle}, we decided to transfer one of our models trained on DSEC-Flow to MVSEC. More specifically, we chose the model trained with dt$_{\text{input}}$ $=$ $0.005$s and $R$ $=$ $20$ 
from \tabref{tab:time}, as a model trained with a short input window on DSEC-Flow is expected to be robust to the slow motion statistics of MVSEC \cite{gehrig2021raft}. We deployed the model at the same frequency as the temporally-upsampled ground truth (i.e., 45 Hz). Results in \tabref{tab:mvsec} show that, by doing this, our model outperforms the great majority of methods in terms of the EPE (even some SL methods trained on this dataset), and is only surpassed by the current state-of-the-art E-RAFT \cite{gehrig2021raft} and TMA \cite{liu2023tma} architectures. Besides reaffirming the high quality of the produced optical flow estimates, these results also confirm the generalizability of our method. For completeness, the results on the indoor evaluation sequences from MVSEC are provided in the supplementary material.

\section{Limitations}
\label{sec:limitations}

The self-supervised method for event-based optical flow presented in this work, while demonstrating highly accurate and promising results, is not without limitations. Two critical challenges that need to be acknowledged are the brightness constancy assumption and the aperture problem. Firstly, the contrast maximization framework \cite{gallego2018unifying, gallego2019focus} assumes constant illumination, leading our models to face difficulty in learning from events that are not due to motion in the image space but that arise from changes in illumination. Since this limitation is inherent to contrast maximization, it extends to other approaches based on the same principle \cite{paredes2021back, hagenaars2021self, shiba2022secrets}. Due to this assumption, we excluded sequences recorded at night from our training dataset (see supplementary material). Secondly, akin to many other optical flow methods, our approach is susceptible to the aperture problem. This indicates that only motion components normal to the orientation of an edge in the image space, also known as normal optical flow, can be reliably resolved \cite{de2020neural}. Consequently, the proposed method might face challenges in accurately determining the true motion direction in certain ambiguous scenarios. The regularizing effect of the iterative event warping (see \secref{sec:nlew}) and the multiple spatial scales at which dense optical flow is estimated in our architecture (see \secref{sec:arch}) are mechanisms in our proposed solution that collectively strive to counteract the aperture problem's influence.

\section{Conclusion}
\label{sec:conclusion}

In this paper, we presented the first learning-based approach to event-based optical flow estimation that is scalable to high inference frequencies while being able to accurately capture the true trajectory of scene points over time. The proposed pipeline is designed around a continuously-running stateful model that sequentially processes fine discrete partitions of the input event stream while integrating spatiotemporal information. We train this model through a novel, self-supervised, contrast maximization framework (i.e., event deblurring for supervision) that is characterized by an iterative event warping module and a multi-timescale loss function that add robustness and improve the accuracy of the predicted optical flow maps. We demonstrated the effectiveness of our approach on multiple datasets, where our models outperform the self-supervised and model-based baselines by large margins. Future research should look into how to learn to better combine the information from multiple timescales, as well as into the design of lightweight architectures that can keep up with real-time constraints.\newpage

We believe that the proposed approach opens up avenues for future research, especially in the field of neuromorphic computing. Spiking networks running on neuromorphic hardware have the potential of exploiting the main benefits of event cameras, but for that they need to process the input events shortly after they arrive.
Our proposed framework is a step toward this objective, as it enables the estimation of optical flow in a close to continuous manner, with all the integration of information happening within the model itself.

{\small
	\bibliographystyle{ieee_fullname}

}


 \clearpage
 \appendix
 
\section{DSEC-Flow: Sequence selection}\label{app:dsec}

The contrast maximization framework for motion compensation assumes constant illumination \cite{gallego2018unifying, gallego2019focus}. Under this assumption, all the events captured with an event camera are generated by the apparent motion of objects in the image space. However, the constant illumination assumption is often violated in sequences recorded at night because the main source of light in these environments comes from flashing, artificial lights (e.g., street lamps). In addition, the signal-to-noise ratio of these sensors decreases under low light conditions, which means that a large percentage of the captured events are not triggered by motion but by sensor noise \cite{gallego2019event}. For these reasons, we remove any sequence recorded at night from the original DSEC-Flow training dataset \cite{gehrig2021dsec, gehrig2021raft}. Specifically, we use all the sequences in the training dataset, from beginning to end, except for variants of those listed in \tabref{tab:notrain}.

\renewcommand{\thetable}{S1}
\begin{table}[!h]
	\centering
	\resizebox{0.575\linewidth}{!}{%
		{\renewcommand{\arraystretch}{1.1} 
			\begin{tabular}{cc}
				\thickhline
				\thickhline
				zurich\_city\_00\_* & zurich\_city\_01\_* \\
				zurich\_city\_09\_* & zurich\_city\_10\_* \\
				\thickhline
				\thickhline
	\end{tabular}}}
	\caption{Sequences from the DSEC-Flow training dataset \cite{gehrig2021dsec, gehrig2021raft} that were not used for the training of our models.}
	\label{tab:notrain}
\end{table}

\section{Additional results}

\subsection{Breakdown of DSEC-Flow results}

A breakdown of the quantitative results on DSEC-Flow can be found in \tabref{tab:breakdown}. Additionally, \figref{fig:DSEC_iwe} shows the IWEs that correspond to the input events and pixel displacement predictions in Fig. 7 and discussed in Section 4.2.

\renewcommand{\thefigure}{S1}
\begin{figure*}[t]
	\centering
	\includegraphics[width=0.725\textwidth]{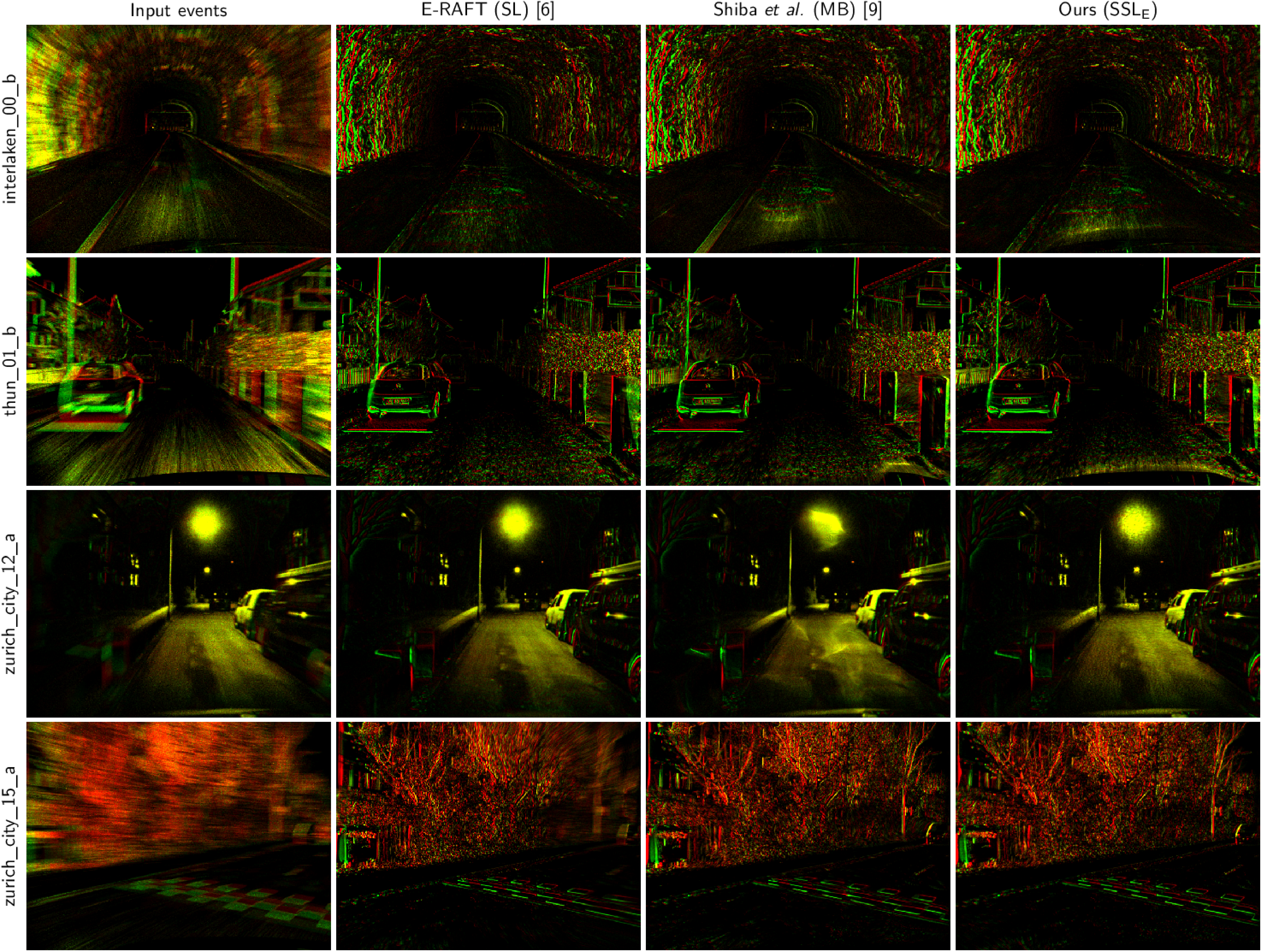}
	\vspace{-5pt}
	\caption{IWEs corresponding to the qualitative comparison of our method with the state-of-the-art E-RAFT architecture \cite{gehrig2021raft} and the model-based approach from Shiba \textit{et al}.\ \cite{shiba2022secrets} on sequences from the test partition of the DSEC-Flow dataset \cite{gehrig2021raft} (see Fig. 7).}
	\label{fig:DSEC_iwe}
\end{figure*}

\renewcommand{\thetable}{S2}
\begin{table}[!t]
	\centering
	\resizebox{0.45\linewidth}{!}{%
		{\renewcommand{\arraystretch}{1.1} 
			\begin{tabular}{lcc}
				\thickhline
				\thickhline
				& EPE$\downarrow$ & $\%_{3\text{PE}}$$\downarrow$\\\thickhline
				dt $=0.1$s$^{*}$ & 3.48 & 34.72 \\
				\hdashline
				dt $=0.05$s$^{*}$ & 3.24 & 32.45 \\
				dt $=0.01$s$^{*}$ & 15.85 & 90.52 \\
				dt $=0.005$s$^{*}$ & 28.45 & 97.27 \\
				dt $=0.002$s$^{*}$ & 15.28 & 94.79\\
				\hdashline			
				dt $=0.05$s & 3.09 & 27.36 \\
				dt $=0.01$s & \textbf{2.33} & \textbf{17.77} \\
				dt $=0.005$s & \underline{2.34} & \underline{17.92} \\
				dt $=0.002$s & 2.66 & 21.83 \\
				\thickhline
				\thickhline
	\end{tabular}}}
	\caption{Quantitative evaluation of the impact of sequential processing on DSEC-Flow \cite{gehrig2021raft}. Best in bold, runner up underlined. $\ast$: Non-recurrent, volumetric event representation with 10 bins.}
	\label{tab:sequential}
\end{table}

\renewcommand{\thetable}{S3}
\begin{table*}[!t]
	\centering
	\resizebox{0.945\linewidth}{!}{%
		{\renewcommand{\arraystretch}{1.1} 
			\begin{tabular}{lccccccccccccccccccc}
				\thickhline
				\thickhline
				& \multicolumn{4}{c}{All} & & \multicolumn{4}{c}{interlaken\_00\_b} & & \multicolumn{4}{c}{interlaken\_01\_a} & & \multicolumn{4}{c}{thun\_01\_a}\\
				\cline{2-5}\cline{7-10}\cline{12-15}\cline{17-20}
				& EPE$\downarrow$ & $\%_{3\text{PE}}$$\downarrow$& FWL$\uparrow$& RSAT$\downarrow$ & & EPE$\downarrow$ & $\%_{3\text{PE}}$$\downarrow$& FWL$\uparrow$& RSAT$\downarrow$&& EPE$\downarrow$ & $\%_{3\text{PE}}$$\downarrow$& FWL$\uparrow$& RSAT$\downarrow$&& EPE$\downarrow$ & $\%_{3\text{PE}}$$\downarrow$& FWL$\uparrow$& RSAT$\downarrow$\\\thickhline
				\parbox[t]{0mm}{\multirow{5}{*}{\rotatebox[origin=c]{90}{\footnotesize SL}}}
				\hspace{10pt}E-RAFT \cite{gehrig2021raft}& 0.79 & 2.68 & 1.33 & \underline{0.87} &  & 1.39 & 6.19 & 1.42 & 0.91 && 0.90 & 3.91 & 1.56 & 0.85 && 0.65 & 1.87 & 1.30 & \underline{0.88} \\
				\hspace{12.5pt}EV-FlowNet, Gehrig \textit{et al}.\ \cite{gehrig2021raft} & 2.32 & 18.60 &  -&- &  & - & - &  -&- && - & - &  -&- && - & - &  -&- \\
				\hspace{12.5pt}IDNet \cite{wu2022lightweight} & \textbf{0.72} & \textbf{2.04} &  -&- &  & \textbf{1.25} & \textbf{4.35} &  -&- && \textbf{0.77} & \textbf{2.60} &  -&- && \textbf{0.57} & \textbf{1.47} &  -&- \\
				\hspace{12.5pt}TIDNet \cite{wu2022lightweight} & 0.84 & 2.80 &  -&- &  & 1.43 & 6.30 &  -&- && 0.93 & 3.50 &  -&- && 0.73 & 2.60 &  -&- \\
				\hspace{12.5pt}TMA \cite{liu2023tma} & \underline{0.74} & \underline{2.30} &  -&- &  & 1.39 & \underline{5.79} &  -&- && \underline{0.81} & \underline{3.11} &  -&- && 0.62 & 1.61 &  -&- \\
				\hspace{12.5pt}Cuadrado \textit{et al}.\ \cite{cuadrado2023optical} & 1.71 & 10.31 &  -&- &  & 3.07 & 23.51 &  -&- && 1.90 & 14.93 &  -&- && 1.36 & 5.93 &  -&- \\
				\hspace{12.5pt}E-Flowformer \cite{li2023blinkflow} & 0.76 & 2.45 &  -&- &  & \underline{1.38} & 6.05 &  -&- && 0.86 & 3.31 &  -&- && \underline{0.60} & \underline{1.60} &  -&- \\
				\hdashline
				\parbox[t]{0mm}{\multirow{8}{*}{\rotatebox[origin=c]{90}{\footnotesize SSL$_{\text{E}}$}}}
				\hspace{10pt}EV-FlowNet$^{*}$ \cite{zhu2019unsupervised} & 3.86 & 31.45 & 1.30 & \textbf{0.85} &  & 6.32 & 47.95 & 1.46 &\textbf{0.85} && 4.91 & 36.07 & 1.42 & \textbf{0.81} && 2.33 & 20.92 & 1.32 & \textbf{0.85} \\
				\hspace{12.5pt}ConvGRU-EV-FlowNet$^{*}$ \cite{hagenaars2021self} & 4.27 & 33.27 & 1.55 & 0.90 &  & 6.78 & 46.77 & \underline{1.74} & 0.92 && 5.21 & 32.26 & \underline{1.92} & 0.88 && 2.15 & 17.50 & \underline{1.49} & 0.91 \\
				\hspace{12.5pt}dt $=0.01$s, $R = 2$, $S=1$ (Ours) & 9.66 & 86.44 & \textbf{1.91} & 1.07 &  & 9.86 & 87.24 & \textbf{1.89} & 1.08 && 9.33 & 86.70 & \textbf{2.07} & 1.01 && 8.71 & 86.45 & \textbf{1.81} & 1.11 \\
				\hspace{12.5pt}dt $=0.01$s, $R = 5$, $S=1$ (Ours) & 4.05 & 52.22 & \underline{1.58} & 0.97 &  & 5.18 & 61.14 & 1.62 & 0.98 && 3.86 & 53.19 & 1.91 & 0.91 && 3.34 & 44.37 & 1.46 & 0.99 \\
				\hspace{12.5pt}dt $=0.01$s, $R = 10$, $S=1$ (Ours) & 2.33 & 17.77 & 1.26 & 0.88 &  & 3.34 & 25.72 & 1.33 & \underline{0.90} && 2.49 & 19.15 & 1.40 & \underline{0.83} && 1.73 & 10.39 & 1.21 & 0.89 \\
				\hspace{12.5pt}dt $=0.01$s, $R = 20$, $S=1$ (Ours) & 16.63 & 33.67 & 1.06 & 1.10 &  & 30.98 & 46.26 & 0.86 & 1.18 && 7.43 & 33.94 & 1.17 & 0.98 && 14.41 & 21.36 & 0.74 & 1.20 \\
				\hspace{12.5pt}dt $=0.01$s, $R = 10$, $S=3$ (Ours) & 2.82 & 27.09 & 1.37 & 0.92 &  & 3.79 & 37.48 & 1.44 & 0.94 && 2.83 & 28.14 & 1.66 & 0.86 && 2.13 & 18.40 & 1.29 & 0.93 \\
				\hspace{12.5pt}dt $=0.01$s, $R = 20$, $S=4$ (Ours) & 2.73 & 23.73 & 1.24 & 0.90 &  & 3.31 & 29.97 & 1.34 & 0.91 && 2.73 & 25.92 & 1.45 & 0.86 && 1.81 & 13.19 & 1.21 & 0.92 \\
				\hdashline
				\parbox[t]{0mm}{\multirow{1.0}{*}{\rotatebox[origin=c]{90}{\footnotesize MB}}}
				\hspace{12.5pt}Shiba \textit{et al}.\ \cite{shiba2022secrets}& 3.47 & 30.86 & 1.37 & 0.89 & & 5.74 & 38.93 & 1.46 & \underline{0.90}  && 3.74 & 31.37 & 1.63 & 0.88 && 2.12 & 17.68 & 1.32 & 0.89  \\
				\thickhline
				& \multicolumn{4}{c}{thun\_01\_b} & & \multicolumn{4}{c}{zurich\_city\_12\_a} & & \multicolumn{4}{c}{zurich\_city\_14\_c} & & \multicolumn{4}{c}{zurich\_city\_15\_a}\\
				\cline{2-5}\cline{7-10}\cline{12-15}\cline{17-20}
				& EPE$\downarrow$ & $\%_{3\text{PE}}$$\downarrow$& FWL$\uparrow$& RSAT$\downarrow$ & & EPE$\downarrow$ & $\%_{3\text{PE}}$$\downarrow$& FWL$\uparrow$& RSAT$\downarrow$&& EPE$\downarrow$ & $\%_{3\text{PE}}$$\downarrow$& FWL$\uparrow$& RSAT$\downarrow$&& EPE$\downarrow$ & $\%_{3\text{PE}}$$\downarrow$& FWL$\uparrow$& RSAT$\downarrow$\\\thickhline
				\parbox[t]{0mm}{\multirow{5}{*}{\rotatebox[origin=c]{90}{\footnotesize SL}}}
				\hspace{10pt}E-RAFT \cite{gehrig2021raft}& 0.58 & 1.52 & 1.25 & \underline{0.89} &  & 0.61 & 1.06 & 0.91 & \textbf{0.93} && 0.71 & \textbf{1.91} & 1.47 & \underline{0.83} && 0.59 & 1.30 & 1.40 & \underline{0.84} \\
				\hspace{12.5pt}EV-FlowNet, Gehrig \textit{et al}.\ \cite{gehrig2021raft} & - & - &  -&- &  & - & - &  -&- && - & - &  -&- && - & - &  -&- \\
				\hspace{12.5pt}IDNet \cite{wu2022lightweight} & \textbf{0.55} & \underline{1.35} &  -&- &  & 0.60 & 1.16 &  -&- && 0.76 & 2.74 &  -&- && \textbf{0.55} & \textbf{1.02} &  -&- \\
				\hspace{12.5pt}TIDNet \cite{wu2022lightweight} & 0.65 & 1.70 &  -&- &  & 0.67 & 1.30 &  -&- && 0.80 & 4.60 &  -&- && 0.65 & 1.30 &  -&- \\
				\hspace{12.5pt}TMA \cite{liu2023tma} & \textbf{0.55} & \textbf{1.31} &  -&- &  & \textbf{0.57} & \textbf{0.87} &  -&- && \textbf{0.66} & \underline{1.99} &  -&- && \textbf{0.55} & \underline{1.08} &  -&- \\
				\hspace{12.5pt}Cuadrado \textit{et al}.\ \cite{cuadrado2023optical} & 1.41 & 6.38 &  -&- &  & 1.24 & 3.85 &  -&- && 1.58 & 9.96 &  -&- && 1.24 & 4.89 &  -&- \\
				\hspace{12.5pt}E-Flowformer \cite{li2023blinkflow} & \underline{0.57} & 1.50 &  -&- &  & \underline{0.58} & \underline{0.91} &  -&- && \underline{0.67} & 2.09 &  -&- && \underline{0.56} & 1.15 &  -&- \\
				\hdashline
				\parbox[t]{0mm}{\multirow{8}{*}{\rotatebox[origin=c]{90}{\footnotesize SSL$_{\text{E}}$}}}
				\hspace{10pt}EV-FlowNet$^{*}$ \cite{zhu2019unsupervised} & 3.04 & 25.41 & 1.33 & \textbf{0.87} &  & 2.62 & 25.80 & 1.03 & \underline{0.94} && 3.36 & 36.34 & 1.24 & \textbf{0.82} && 2.97 & 25.53 & 1.33 & \textbf{0.82} \\
				\hspace{12.5pt}ConvGRU-EV-FlowNet$^{*}$ \cite{hagenaars2021self} & 3.25 & 25.31 & \underline{1.51} & 0.92 &  & 3.67 & 40.15 & 0.97 & \textbf{0.93} && 3.47 & 40.98 & 1.60 & 0.87 && 3.21 & 27.99 & \underline{1.61} & 0.89 \\
				\hspace{12.5pt}dt $=0.01$s, $R = 2$, $S=1$ (Ours) & 9.38 & 86.68 & \textbf{1.66} & 1.08 &  & 11.54 & 85.35 & \underline{1.40} &1.10 && 10.18 & 86.39 & \textbf{2.50} & 1.03 && 8.54 & 86.30 & \textbf{2.01} & 1.06 \\
				\hspace{12.5pt}dt $=0.01$s, $R = 5$, $S=1$ (Ours) & 3.51 & 47.33 & \textbf{1.66} & 1.00 &  & 4.76 & 51.82 & 1.14 & 0.99 && 4.23 & 57.26 & \underline{1.72} & 0.93 && 3.42 & 50.40 & 1.54 & 0.96 \\
				\hspace{12.5pt}dt $=0.01$s, $R = 10$, $S=1$ (Ours) & 1.66 & 9.34 & 1.25 & 0.91 &  & 2.72 & 26.65 & 1.04 & \underline{0.94} && 2.64 & 23.01 & 1.38 & 0.85 && 1.69 & 9.98 & 1.23 & 0.86 \\
				\hspace{12.5pt}dt $=0.01$s, $R = 20$, $S=1$ (Ours) & 9.09 & 22.53 & 1.09 & 1.08 &  & 27.19 & 44.78 & \textbf{1.49} & 1.17 && 20.06 & 41.65 & 1.16 & 1.05 && 15.97 & 25.16 & 0.90 & 1.05 \\
				\hspace{12.5pt}dt $=0.01$s, $R = 10$, $S=3$ (Ours) & 2.03 & 17.19 & 1.40 & 0.94 &  & 3.53 & 33.77 & 1.08 &0.97 && 2.95 & 32.75 & 1.43 & 0.88 && 2.26 & 21.95 & 1.29 & 0.89 \\
				\hspace{12.5pt}dt $=0.01$s, $R = 20$, $S=4$ (Ours) & 1.85 & 13.71 & 1.21 & 0.93 &  & 4.19 & 35.65 & 0.91 &0.95 && 2.53 & 27.97 & 1.32 & 0.85 && 1.93 & 15.50 & 1.24 & 0.87 \\
				\hdashline
				\parbox[t]{0mm}{\multirow{1.0}{*}{\rotatebox[origin=c]{90}{\footnotesize MB}}}
				\hspace{12.5pt}Shiba \textit{et al}.\ \cite{shiba2022secrets}& 2.48 & 23.56 & 1.28 & \underline{0.89}  & & 3.86 & 43.96 & 1.08&0.95 && 2.72 & 30.53 & 1.44&0.85 && 2.35 & 20.99 & 1.39&0.87 \\
				\thickhline
				\thickhline
				\multicolumn{4}{l}{\hspace{12.5pt}\small $^*$Retrained by us on DSEC-Flow, linear warping.}
\end{tabular}}}
\vspace{-5pt}
\caption{Breakdown of the quantitative evaluation on the DSEC-Flow dataset \cite{gehrig2021raft}. Best in bold, runner up underlined. The results of our best performing model are highlighted in red. SL: supervised learning; SSL$_{\text{F}}$: SSL trained with grayscale images; SSL$_{\text{E}}$; SSL trained with events; MB: model-based methods.}
\vspace{-5pt}
\label{tab:breakdown}
\end{table*}

\renewcommand{\thefigure}{S2}
\begin{figure*}[!t]
	\centering
	\begin{subfigure}[b]{0.475\textwidth}
		\centering
		\includegraphics[width=0.9\textwidth]{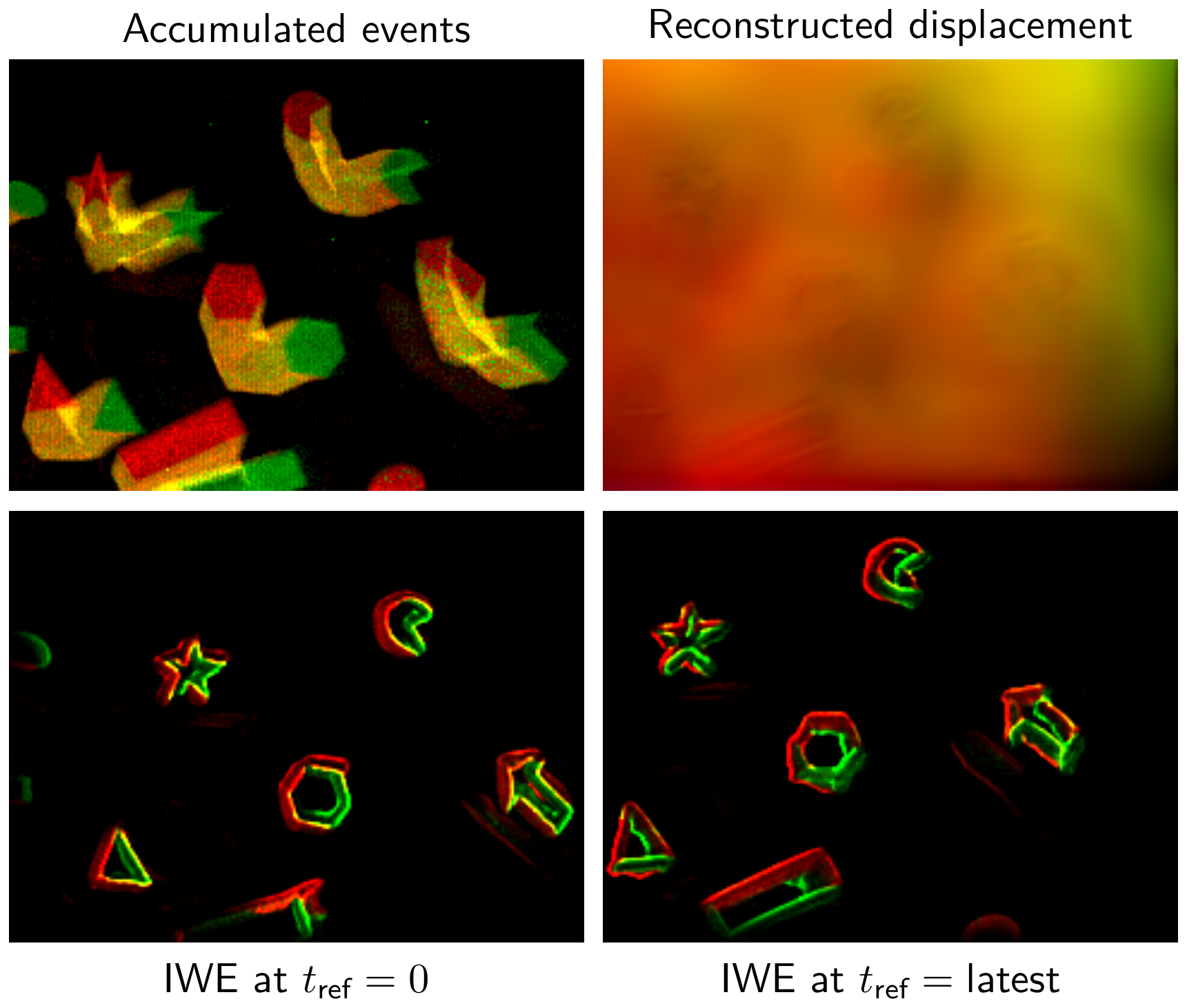}
		\caption{Ours.}
	\end{subfigure}
	\begin{subfigure}[b]{0.475\textwidth}
		\centering
		\includegraphics[width=0.9\textwidth]{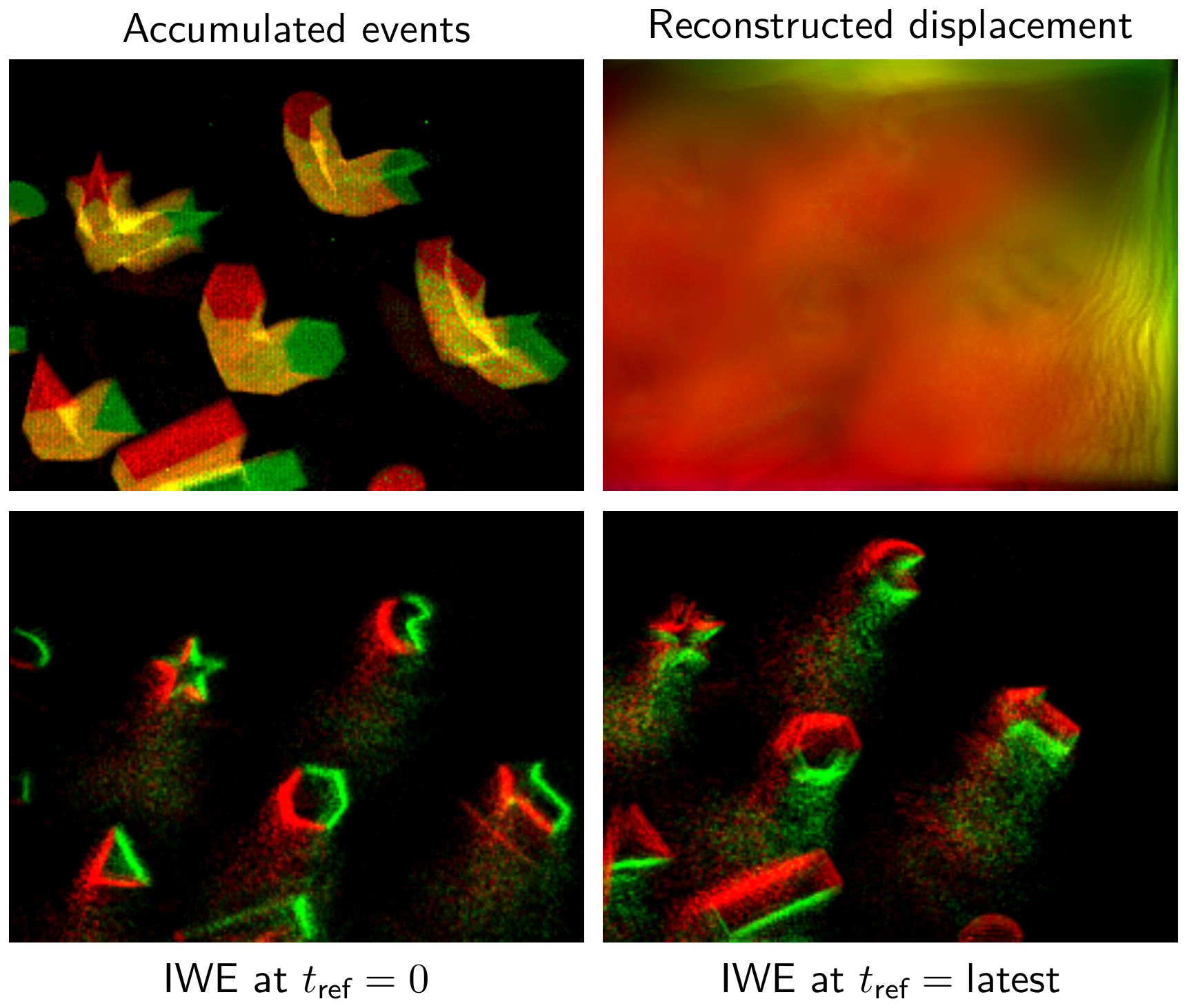}
		\caption{ConvGRU-EV-FlowNet$^{*\dagger}$ \cite{hagenaars2021self}.}
	\end{subfigure}
	\caption{Qualitative results of the ablation study with respect to the type of event warping. Inference settings: dt $=0.01$s, accumulation window of $0.5$s. Both models were trained on the DSEC-Flow dataset \cite{gehrig2021raft}, with dt $=0.01$s, $R = 10$, $S=1$. 
		$\ast$: Retrained by us on DSEC-Flow \cite{gehrig2021raft}. $\dagger$: Without border compensation. The optical flow color coding can be found in Fig. 2 (top).}
\label{fig:ablation_warping}
\end{figure*}


\subsection{Breakdown of MVSEC results}

\figref{fig:MVSEC} shows a qualitative comparison of our self-supervised learning (SSL) method with state-of-the-art techniques on the outdoor\_day1 sequence from MVSEC. As described in Section 4.2, for this evaluation we used all the events in between samples of the temporally-upsampled ground-truth data (provided at 45 Hz) as input for every forward pass. These results support the conclusions derived from Table 3.

\renewcommand{\thefigure}{S3}
\begin{figure*}[t]
\centering
\includegraphics[width=\textwidth]{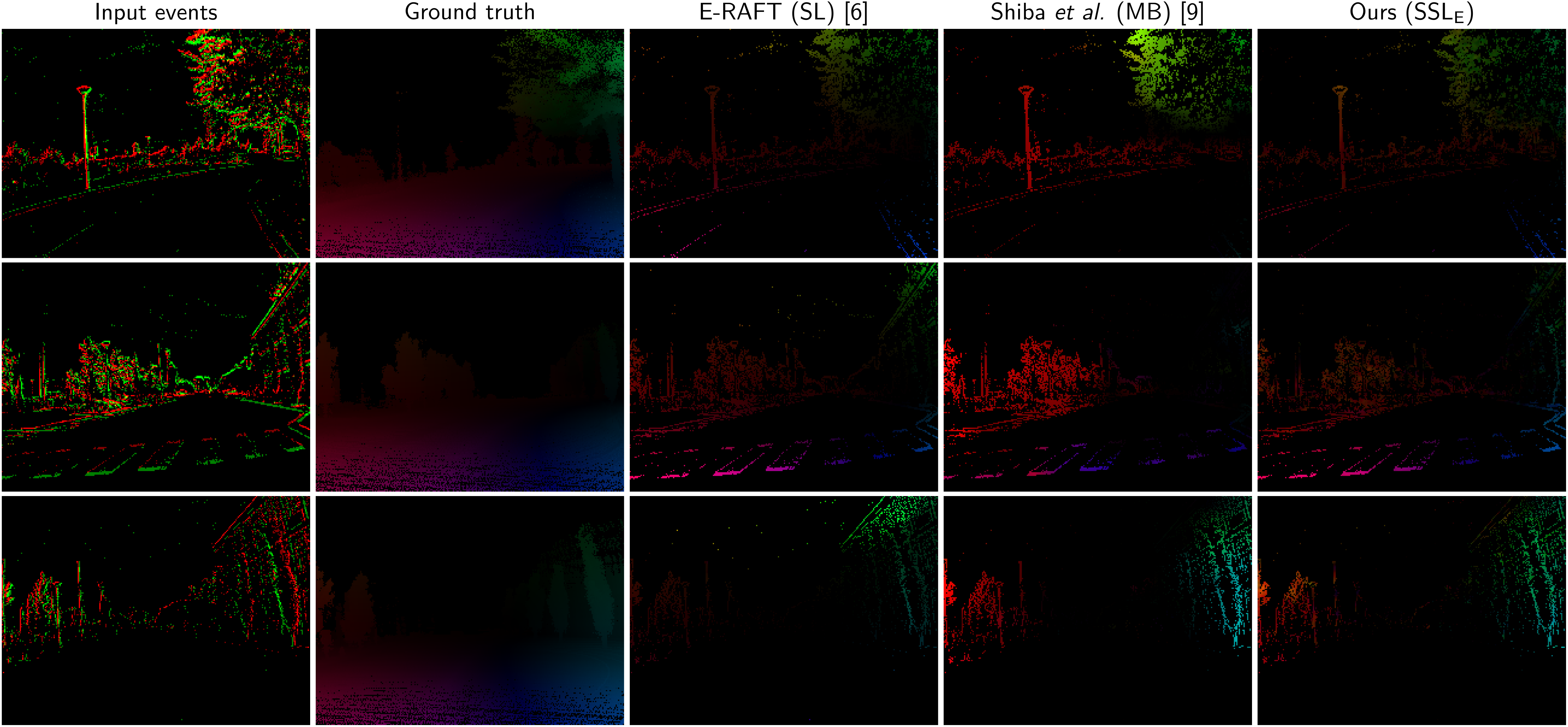}
\caption{Qualitative comparison of our method with the state-of-the-art E-RAFT architecture \cite{gehrig2021raft} and the model-based approach from Shiba \textit{et al}.\ \cite{shiba2022secrets} on the outdoor\_day1 sequence from the MVSEC dataset \cite{zhu2018multivehicle}. Optical flow predictions are masked with the input events to be consistent with the evaluation proposed in \cite{zhu2018ev}. The optical flow color coding can be found in Fig. 2 (top).}
\label{fig:MVSEC}
\end{figure*}

For completeness, a breakdown of the quantitative results on all MVSEC evaluation sequences can be found in Table \ref{tab:mvsec_all}. This includes not just the outdoor\_day1 scenario (as discussed in the main text), but also the three indoor sequences. These indoor sequences present notably different statistics compared to the automotive dataset used for training our models \cite{gehrig2021raft}, as they were recorded with a drone operating in an indoor environment. On these sequences, our transferred model demonstrates (on average) an improvement of $\downarrow\hspace{-3pt}25\%$ in endpoint error (EPE, lower is better, $\downarrow$) compared to the architecturally-equivalent ConvGRU-EV-FlowNet model from Hagenaars \textit{et al.} \cite{hagenaars2021self}, while showing an error increase of $\uparrow\hspace{-3pt}30\%$ compared to Shiba et al. \cite{shiba2022secrets}. However, note that the latter method is not learning-based, so it is not subject to generalization issues besides those inherent to contrast maximization (see main text). Lastly, it's worth noting that the benchmarking of event-based optical flow solutions is shifting from MVSEC \cite{zhu2018multivehicle} to the DSEC dataset \cite{gehrig2021raft} due to calibration issues \cite{gehrig2021raft} and the lack of a standardized training dataset in the former \cite{gehrig2021raft,wu2022lightweight,liu2023tma, cuadrado2023optical, li2023blinkflow, shiba2022secrets, hagenaars2021self, stoffregen2020train}.

\subsection{Impact of sequential processing}

Here, we study the impact of the proposed contrast maximization framework for sequential event-based optical flow estimation (i.e., short input partitions, longer training partitions; see Section 3) and compare it to the non-sequential pipeline from Zhu \textit{et al}.\ \cite{zhu2019unsupervised} (i.e., input and training partitions are of the same length). To do this, we trained multiple models on DSEC-Flow with different dt$_{\text{input}}$, but with $\text{dt}_{\text{train}}$$=0.1$s for the sequential models and $\text{dt}_{\text{train}}$$=\text{dt}_{\text{input}}$ for the non-sequential. Quantitative results in \tabref{tab:sequential} confirm the claims made in Section 3.1 about the fact that, for contrast maximization to be a robust supervisory signal, the training event partition used for the computation of the supervisory signal needs to contain enough motion information (i.e., blur) so it can be compensated for. As shown, non-sequential models converge to worse solutions the shorter the input window. On the other hand, our sequential pipeline allows us to shorten the input window without compromising the performance, as discussed in Section 4.2.

\renewcommand{\thetable}{S4}
\begin{table*}[!t]
\vspace{-5pt}
\centering
\resizebox{0.74\linewidth}{!}{%
{\renewcommand{\arraystretch}{1.1} 
	\begin{tabular}{lcccccccccccc}
		\thickhline
		\thickhline
		\multirow{2}{*}{} & \multicolumn{2}{c}{outdoor\_day1} && \multicolumn{2}{c}{indoor\_flying1} && \multicolumn{2}{c}{indoor\_flying2} && \multicolumn{2}{c}{indoor\_flying3} \\\cline{2-3}\cline{5-6}\cline{8-9}\cline{11-13}
		& EPE$\downarrow$& $\%_{\text{3PE}}$$\downarrow$&& EPE$\downarrow$& $\%_{\text{3PE}}$$\downarrow$&& EPE$\downarrow$& $\%_{\text{3PE}}$$\downarrow$&& EPE$\downarrow$& $\%_{\text{3PE}}$$\downarrow$\\\thickhline
		\parbox[t]{0mm}{\multirow{5}{*}{\rotatebox[origin=c]{90}{SL}}}
		\hspace{10pt}EV-FlowNet+ \cite{stoffregen2020train} & 0.68 & 0.99 && 0.56 & 1.00 && \underline{0.66} & \underline{1.00} && \underline{0.59} & \underline{1.00}\\
		\hspace{12.5pt}E-RAFT \cite{gehrig2021raft} & \textbf{0.24} & 1.70 && - & - && - & - && - & - \\
		\hspace{12.5pt}EV-FlowNet \cite{gehrig2021raft} & 0.31 & \textbf{0.00} && - & - && - & - && - & - \\
		\hspace{12.5pt}TMA \cite{liu2023tma} & \underline{0.25} & 0.07 && 1.06 & 3.63 && 1.81 & 27.29 && 1.58 & 23.26 \\
		\hspace{12.5pt}Cuadrado \textit{et al}.\ \cite{cuadrado2023optical}& 0.85 & -&& 0.58 & - && 0.72 & - && 0.67 & - \\
		\hdashline
		\parbox[t]{0mm}{\multirow{2}{*}{\rotatebox[origin=c]{90}{SSL$_{\text{F}}$}}}
		\hspace{10pt}EV-FlowNet \cite{zhu2018ev} & 0.49 & 0.20 && 1.03 & 2.20 && 1.72 & 15.1 && 1.53 & 11.9\\
		\hspace{12.5pt}Ziluo \textit{et al}.\ \cite{ding2022spatio} & 0.42 & \textbf{0.00} && 0.57 & 0.10 && 0.79 & 1.60 && 0.72 & 1.30 \\
		\hdashline
		\parbox[t]{0mm}{\multirow{5}{*}{\rotatebox[origin=c]{90}{SSL$_{\text{E}}$}}}
		\hspace{10pt}EV-FlowNet \cite{zhu2019unsupervised} & 0.32 & \textbf{0.00} && 0.58 & \textbf{0.00} && 1.02 & 4.00 && 0.87 & 3.00 \\
		\hspace{12.5pt}EV-FlowNet \cite{paredes2021back} & 0.92 & 5.40 && 0.79 & 1.20 && 1.40 & 10.9 && 1.18 & 7.40 \\
		\hspace{12.5pt}EV-FlowNet \cite{shiba2022secrets} & 0.36 & 0.09 && - & - && - & - && - & -\\
		\hspace{12.5pt}ConvGRU-EV-FlowNet \cite{hagenaars2021self} & 0.47 & 0.25 && 0.60 & 0.51 && 1.17 & 8.06 && 0.93 & 5.64 \\
		\hspace{12.5pt}Ours $dt = 0.005$ & 0.27 & \underline{0.05} && \underline{0.44} & \textbf{0.00} && 0.88 & 4.51 && 0.70 & 2.41 \\  
		\hdashline
		\parbox[t]{0mm}{\multirow{3}{*}{\rotatebox[origin=c]{90}{MB}}}
		\hspace{10pt}Akolkar \textit{et al}.\ \cite{akolkar2020real} & 2.75 & - && 1.52 & - && 1.59 & - && 1.89 & - \\
		\hspace{12.5pt}Brebion \textit{et al}.\ \cite{brebion2021real} & 0.53 & 0.20 && 0.52 & 0.10 && 0.98 & 5.50 && 0.71 & 2.10\\
		\hspace{12.5pt}Shiba \textit{et al}.\ \cite{shiba2022secrets} & 0.30 & 0.11 && \textbf{0.42} & \underline{0.09} && \textbf{0.60} & \textbf{0.59} && \textbf{0.50} & \textbf{0.29}\\
		\thickhline
		\thickhline
\end{tabular}}}
\caption{Quantitative evaluation on all MVSEC sequences \cite{zhu2018multivehicle}. Best in bold, runner up underlined. SL: supervised learning; SSL$_{\text{F}}$: SSL trained with grayscale images; SSL$_{\text{E}}$; SSL trained with events; MB: model-based methods.}
\label{tab:mvsec_all}
\end{table*}

\subsection{Linear vs. iterative event warping}

Here, we examine the effect of the type of event warping (linear \cite{hagenaars2021self} vs.\ iterative) on the performance of the sequential, stateful architecture introduced in Section 3.4 when it is trained on the DSEC-Flow dataset. To do this, we trained four models in total: two variants (with and without image border compensation, see \secref{sec:ablation_border}) of ConvGRU-EV-FlowNet \cite{hagenaars2021self}, which is trained with linear warping; and another two variants of the \textit{same architecture}, but trained with the proposed iterative warping module. Quantitative and qualitative results are presented in \tabref{tab:ablation_border} and \figref{fig:ablation_border}, respectively. In both cases (with and without image-border compensation), the models trained with iterative warping (i.e., ours) outperform those trained with linear warping (EPE dropped by $28\%$ without compensation, and $62\%$ with it), despite using the same architecture. This is expected, as the iterative warping module is able to better capture the trajectory of scene points over time, as explained in Section 3.2. The impact of the image-border compensation mechanism is presented and discussed in \secref{sec:ablation_border}.

To support the arguments presented in Section 3.2 and Fig. 2 regarding the limitations of linear warping, we also conducted an experiment in which we deployed models trained on DSEC-Flow with linear and iterative warping on a sequence from the Event Camera Dataset \cite{mueggler2017event} with strong nonlinearities in the trajectories of scene points. Note that this sequence, known as shapes$\_$6dof, was recorded with a different event camera and that its statistics are significantly different from those of DSEC-Flow (i.e., hand-held camera looking at a planar scene \cite{mueggler2017event} vs. automotive scenario \cite{gehrig2021dsec}). Qualitative results are presented in \figref{fig:ablation_warping}. In addition to showing that the models generalize (to some extent) to this new sequence, these results demonstrate that only the models trained with iterative event warping are able to produce sharp IWEs at multiple reference times.

\renewcommand{\thefigure}{S4}
\begin{figure*}[!h]
\centering
\includegraphics[width=0.925\textwidth]{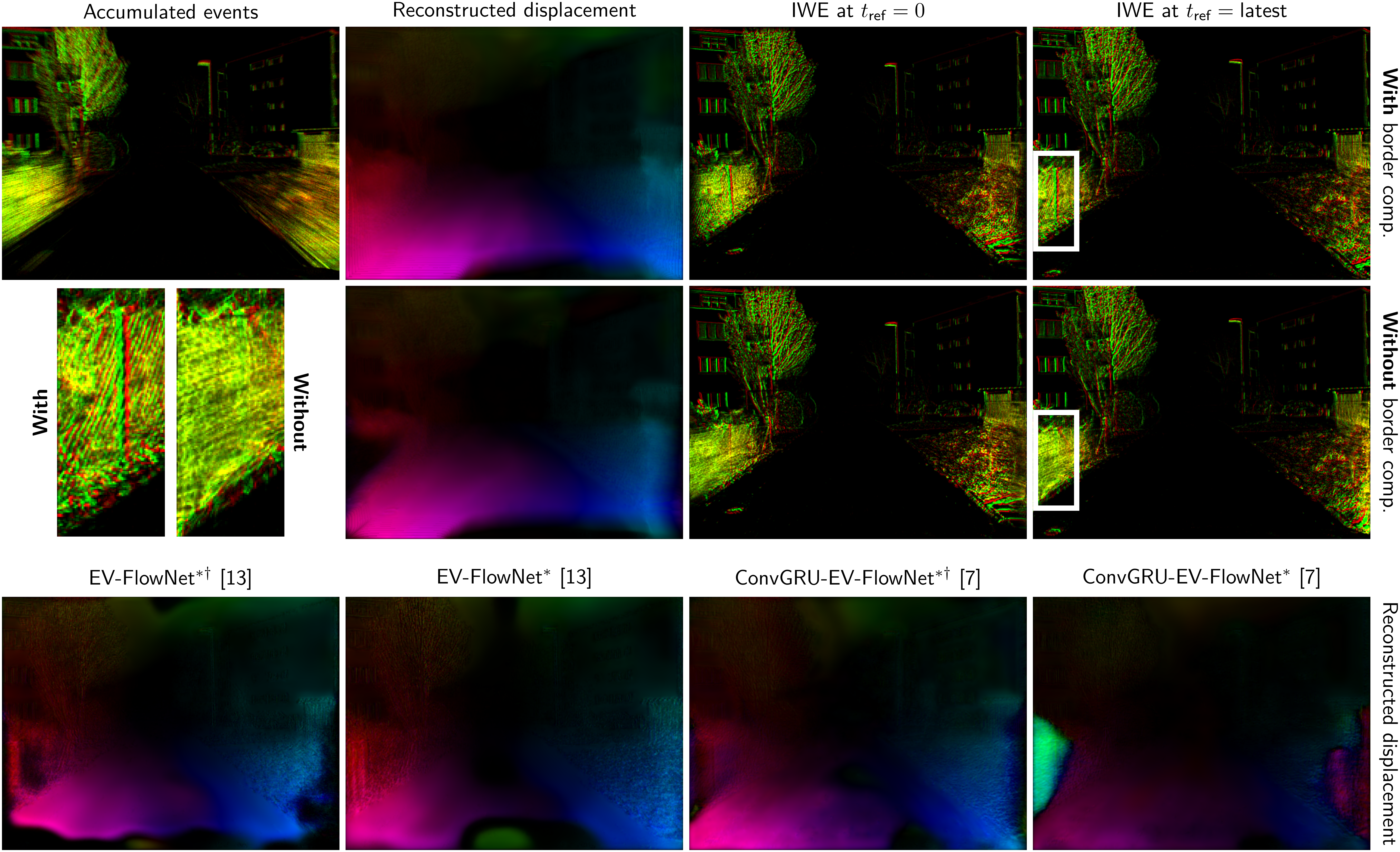}
\caption{Qualitative results of the ablation study on the DSEC-Flow dataset \cite{gehrig2021raft} with respect to the effectiveness of the proposed event warping module and image-border compensation mechanism. \textit{Top}: Models trained with the proposed SSL framework (dt $=0.01$s, $R = 10$, $S=1$). \textit{Bottom}: Literature methods EV-FlowNet \cite{zhu2019unsupervised} and ConvGRU-EV-FlowNet \cite{hagenaars2021self}. $\ast$: Retrained by us on DSEC-Flow. $\dagger$: Without border compensation. The optical flow color coding can be found in Fig. 2 (top).}
\label{fig:ablation_border}
\end{figure*}

\renewcommand{\thefigure}{S5}
\begin{figure*}[t]
	\centering
	\includegraphics[width=0.925\textwidth]{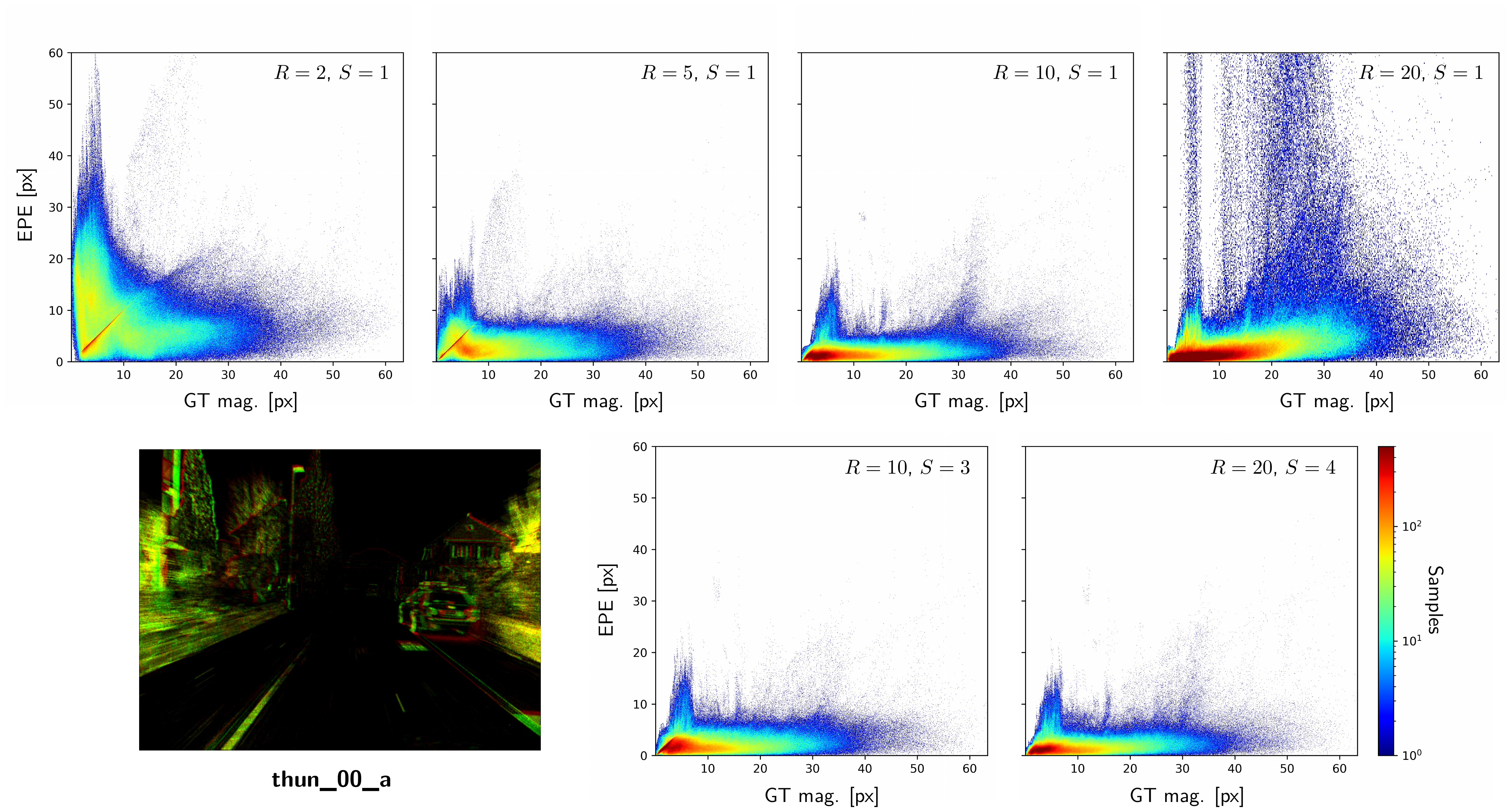}
	\caption{Distribution of the EPE of our models in Table 1 as a function of the ground truth magnitude in the thun\_00\_a sequence from DSEC-Flow \cite{gehrig2021raft}. For this experiment, and as in Table 1, all models were trained and deployed with $\text{dt}_{\text{input}}=0.01$s.}
	\label{fig:error_hist}
\end{figure*}

\subsection{Optical flow at the image borders}\label{sec:ablation_border}

\renewcommand{\thetable}{S5}
\begin{table}[!t]
\vspace{-5pt}
\centering
\resizebox{0.75\linewidth}{!}{%
{\renewcommand{\arraystretch}{1.1} 
	\begin{tabular}{lcccc}
		\thickhline
		\thickhline
		& EPE$\downarrow$ & $\%_{3\text{PE}}$$\downarrow$\\\thickhline
		EV-FlowNet$^{*\dagger}$ \cite{zhu2019unsupervised} & 3.86 & 31.45\\ 
		EV-FlowNet$^{*}$ \cite{zhu2019unsupervised} & 3.48 & 34.72\\
		\hdashline
		ConvGRU-EV-FlowNet$^{*\dagger}$ \cite{hagenaars2021self} & 4.27 & 33.27\\
		ConvGRU-EV-FlowNet$^{*}$ \cite{hagenaars2021self} & 6.09 & 36.36\\
		\hdashline
		dt $=0.01$s, $R = 10$, $S=1^{\dagger}$ (Ours) & 3.08 & 21.38\\
		dt $=0.01$s, $R = 10$, $S=1$ (Ours) & 2.33 & 17.77\\
		\thickhline
		\thickhline
		\multicolumn{4}{l}{\small $^*$Retrained by us on DSEC-Flow, linear warping.}\\
	\vspace{-15pt}\\
	\multicolumn{4}{l}{\small $^\dagger$Without border compensation.}
\end{tabular}}}
\caption{Quantitative results of the ablation study on the DSEC-Flow dataset \cite{gehrig2021raft} with respect to the effectiveness of the proposed warping module and image-border compensation mechanism.}
\label{tab:ablation_border}
\end{table}

As discussed in Section 3.2, for a given temporal scale, we mask the events that are transported outside the image space at any time during the warping process from the computation of the loss to prevent learning incorrect optical flow at the image borders. Here we study the impact of this masking mechanism on the performance of not only the proposed SSL framework but also of two other literature methods: EV-FlowNet \cite{zhu2019unsupervised} and ConvGRU-EV-FlowNet \cite{hagenaars2021self}. For this experiment, we trained two versions of each model, one with and one without the proposed image-border compensation technique, on the DSEC-Flow dataset.
Note that EV-FlowNet is a stateless model trained with a volumetric event representation with 10 bins, and hence processes all the input events in between ground-truth samples at once. 

Quantitative and qualitative results are presented in \tabref{tab:ablation_border} and \figref{fig:ablation_border}, respectively. These results highlight that, for both EV-FlowNet and our model, adding the proposed image-border compensation improves performance (EPE dropped by $10\%$ and $24\%$, respectively). However, the performance degraded when adding it to the training pipeline of ConvGRU-EV-FlowNet (EPE went up by $43\%$). We believe that the reason for this drop in performance is the event warping method used during training. While the proposed iterative warping allows for the error to propagate through all the pixels covered in the warping process, the linear warping used to train ConvGRU-EV-FlowNet only propagates the error through pixels with input events \cite{hagenaars2021self}. Therefore, if events are removed from the computation of the loss, the error is not propagated through the corresponding pixels, and then the spatial coherence of the resulting optical flow maps degrades. Despite sharing the same warping methodology, this is less of an issue for EV-FlowNet since it processes the events from longer temporal windows in a single forward pass, producing a single optical flow map per loss. The longer this window, the more likely it is that a pixel contains events triggered by multiple moving objects (i.e., reflected as events with different timestamps), and hence the higher the probability that the error is propagated through that pixel.

\subsection{Visualizing the endpoint error}

To support the hypothesis in Section 3.3 that the length of the training partition $R$ has a significant impact on the quality of the training, here we study the distribution of the EPE of our models in Table 1 as a function of the ground truth optical flow magnitude in the thun\_00\_a\footnote{Note that, since ground truth is required for this experiment, this sequence belongs to the training partition of DSEC-Flow \cite{gehrig2021raft}. Consequently, this means that our models have had access to a randomly cropped version of it during training.} sequence from DSEC-Flow \cite{gehrig2021raft}. The error distributions are shown in \figref{fig:error_hist} and confirm the conclusions derived from Table 1 in Section 4.2. Models trained with short training partitions (i.e., $R\in[2, 5]$) converge to solutions that are less accurate (i.e., high EPE) for low ground truth magnitudes, while long partitions (i.e., $R\geq 20$) do the same but for high ground truth magnitudes. The proposed multi-timescale approach (i.e., $S>1$) to contrast maximization alleviates this issue and allows for the training of models that are accurate for all ground truth magnitudes without having to fine-tune the length of the training partition. As shown in this figure, the error distribution of the $S>1$ models closely resembles that of our best performing solution: $R=10$, $S=1$.


\end{document}